\documentclass[10pt,twocolumn,letterpaper]{article}

\usepackage[pagenumbers]{cvpr} 

\usepackage{multirow}

\definecolor{cvprblue}{rgb}{0.21,0.49,0.74}
\usepackage[pagebackref,breaklinks,colorlinks,allcolors=cvprblue]{hyperref}
\usepackage[accsupp]{axessibility}  

\def\paperID{8588} 
\def\confName{CVPR}
\def\confYear{2025}

\title{Feature4X: Bridging Any Monocular Video to 4D Agentic AI \\ with Versatile Gaussian Feature Fields}

\author{Shijie Zhou$^1$\thanks{Equal contribution.} \quad Hui Ren$^2$\footnotemark[1] \quad Yijia Weng$^3$ \quad  
Shuwang Zhang$^1$ \quad Zhen Wang$^1$ \quad Dejia Xu$^4$ \\ \quad Zhiwen Fan$^4$ \quad Suya You$^5$ \quad Zhangyang Wang$^4$ \quad Leonidas Guibas$^3$ \quad Achuta Kadambi$^1$\\
\normalsize{$^1$UCLA \quad
$^2$MIT \quad
$^3$Stanford \quad
$^4$UT Austin \quad
$^5$DEVCOM ARL}\\
\url{https://feature4x.github.io/}
}

\begin{document}
\twocolumn[{%
\renewcommand\twocolumn[1][]{#1}%
\maketitle
    \captionsetup{type=figure}
    \vspace{-5.5mm}
    \includegraphics[width=\textwidth]{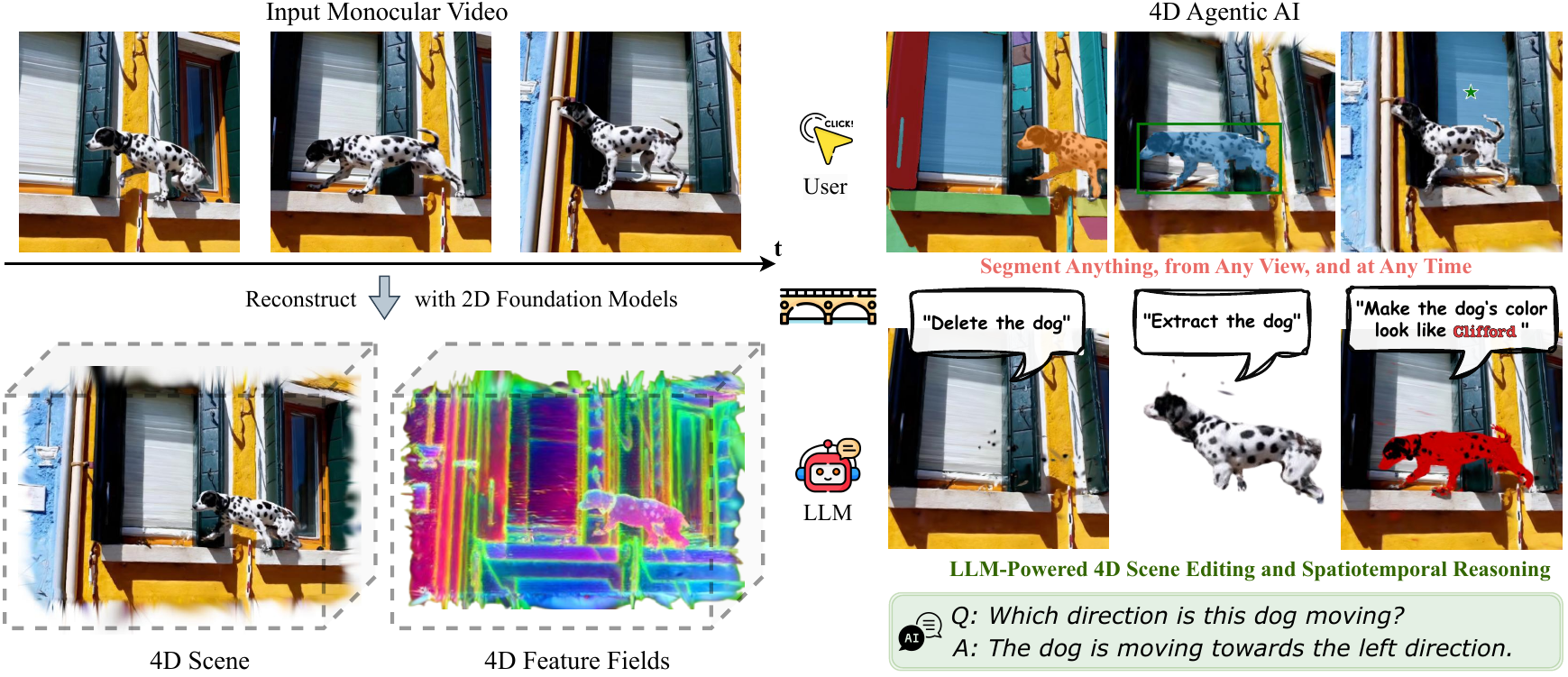} 
    \vspace{-5mm}
    \hfill\caption{\textbf{Feature4X: Building 4D Interactive Scenes with Agentic AI from Monocular Videos.} By dynamically distilling model-conditioned features and integrating 2D foundation models with LLMs in feedback loops, Feature4X enables multimodal tasks across 2D, 3D, and 4D with high-level language inputs or direct user interactions, including (but not limited to) segmentation, scene editing, and VQA across novel views and all time steps, unlocking new possibilities for 4D agentic AI.}
    \label{fig:teaser}
    \hfill \vspace{0mm}
}]
{
\renewcommand{\thefootnote}{\fnsymbol{footnote}}
\footnotetext[1]{Equal contribution.}
}

\begin{abstract}
Recent advancements in 2D and multimodal models have achieved remarkable success by leveraging large-scale training on extensive datasets. However, extending these achievements to enable free-form interactions and high-level semantic operations with complex 3D/4D scenes remains challenging. This difficulty stems from the limited availability of large-scale, annotated 3D/4D or multi-view datasets, which are crucial for generalizable vision and language tasks such as open-vocabulary and prompt-based segmentation, language-guided editing, and visual question answering (VQA). In this paper, we introduce Feature4X, a universal framework designed to extend any functionality from 2D vision foundation model into the 4D realm, using only monocular video input, which is widely available from user-generated content. The ``X" in Feature4X represents its versatility, enabling any task through adaptable, model-conditioned 4D feature field distillation. At the core of our framework is a dynamic optimization strategy that unifies multiple model capabilities into a single representation. Additionally, to the best of our knowledge, Feature4X is the first method to distill and lift the features of video foundation models (e.g., SAM2, InternVideo2) into an explicit 4D feature field using Gaussian Splatting. Our experiments showcase novel view segment anything, geometric and appearance scene editing, and free-form VQA across all time steps, empowered by LLMs in feedback loops. These advancements broaden the scope of agentic AI applications by providing a foundation for scalable, contextually and spatiotemporally aware systems capable of immersive dynamic 4D scene interaction. 

\end{abstract}    
\section{Introduction}
The rapid evolution of 2D vision and multimodal models has been fueled by access to massive curated datasets with rich annotations, alongside breakthroughs spanning multiple domains. These developments have led to remarkable progress in various tasks, including image segmentation~\cite{radford2021learning}, image editing~\cite{brooks2022instructpix2pix}, promptable semantic segmentation~\cite{ravi2024sam,kirillov2023segment}, and Visual Question Answering (VQA)~\cite{wang2024internvideo2,li2023videochat}. Despite these advances, the processing and interpretation of dynamic 3D data—crucial for applications such as autonomous driving, robotics, and 3D asset creation—still lag significantly behind the development of versatile and robust 2D foundation models. A fundamental challenge in 3D vision lies in processing multi-view images and widely available monocular videos, which often suffer from limited camera information and a scarcity of well-annotated, fine-grained per-frame datasets. To bridge this gap, we propose to leverage the latent features of well-trained representations from 2D and multimodal domains~\cite{radford2021learning,ravi2024sam,kirillov2023segment,wang2024internvideo2,li2023videochat} and adaptively lift them into higher-dimensional feature fields (such as 3D and 4D) in a scalable and efficient manner, enabling the direct transfer of 2D functionalities into 4D with minimal data annotation and training overhead.

Constructing 4D feature fields from casually captured in-the-wild videos is significantly challenging. Existing approaches for static 3D feature fields~\cite{yu2024language,shi2024language,zhou2024feature,qin2024langsplat,liu2024splatraj,zuo2024fmgs,lee2024rethinking,kobayashi2022decomposing,kerr2023lerf,tschernezki22neural,fan2024large} are not directly applicable to 4D scenarios for three primary reasons. First, these methods typically rely on well-calibrated, multi-view input images with precise camera poses, which are difficult to obtain from casual videos. Second, extending static feature fields to include an additional temporal dimension results in substantial memory demands, leading to unstable and costly optimization. Third, previous approaches are limited to task-specific feature fields, necessitating full-parameter re-training for adaptation to new tasks.

In this paper, we set an ambitious goal: to parse in-the-wild monocular videos into a unified dynamic 4D representation that not only reconstructs accurate appearance and geometry but also seamlessly integrates advanced functionalities from a range of 2D foundation models. To tackle the aforementioned challenges, we propose enhancing the dynamic 3D Gaussian Splatting-based 4D scene representation~\cite{kerbl20233d,luiten2023dynamic} with a unified latent feature capable of distilling diverse 2D foundation features, achieving both flexibility and efficiency. However, directly lifting 2D feature maps to dense, dynamic 3D Gaussians incurs significant computational costs and fails to scale with the spatial and temporal complexity of the scene. Inspired by recent advancements in Gaussian-based 4D reconstruction~\cite{lei2024mosca,wang2024shape}, we propose to leverage the smooth, compact nature of underlying scene semantics and represent the dense 4D feature field using a sparse set of base or ``Scaffold" features, enabling efficient representation and scalable adaptation to various downstream tasks.

Our pipeline is fully end-to-end differentiable, supervised by ground truth color and feature maps exported from 2D vision foundation models. Its general and model-agnostic design supports a wide range of vision tasks, spanning 2D (semantic and promptable segmentation), 3D (scene editing), and 4D (spatial and temporal VQA). 
Furthermore, our compact and versatile 4D feature field serves as a bridge, seamlessly integrating these tasks with LLMs in a continuous feedback loop. This integration enables intuitive and efficient execution through free-form, high-level natural language inputs or direct user interactions.
By unifying diverse vision tasks, our framework—seamlessly integrated with LLMs—paves the way for the development of advanced, contextually and spatiotemporally aware 4D agentic AI systems. 

We summarize our \textbf{contributions} as follows:
\begin{itemize}
    \item We propose a general 4D feature field distillation technique using Gaussian Splatting to build an interactive 4D scene and lift the functionalities of any 2D vision (image/video) foundation models into the 4D realm, relying solely on monocular video input.
    \item We introduce the first compact and versatile 4D Gaussian feature field representation, which leverages the smooth, low-rank nature of scene semantics by modeling the dense 4D feature field as interpolations of a sparse set of base features.
    \item We develop an LLM-powered agentic AI capable of interpreting natural language prompts, dynamically adjusting configuration parameters, and iteratively refining results through trial and feedback, enabling intelligent 4D scene interaction and manipulation.
\end{itemize}

\begin{figure*}[t]
  \centering
   \includegraphics[width=\linewidth]{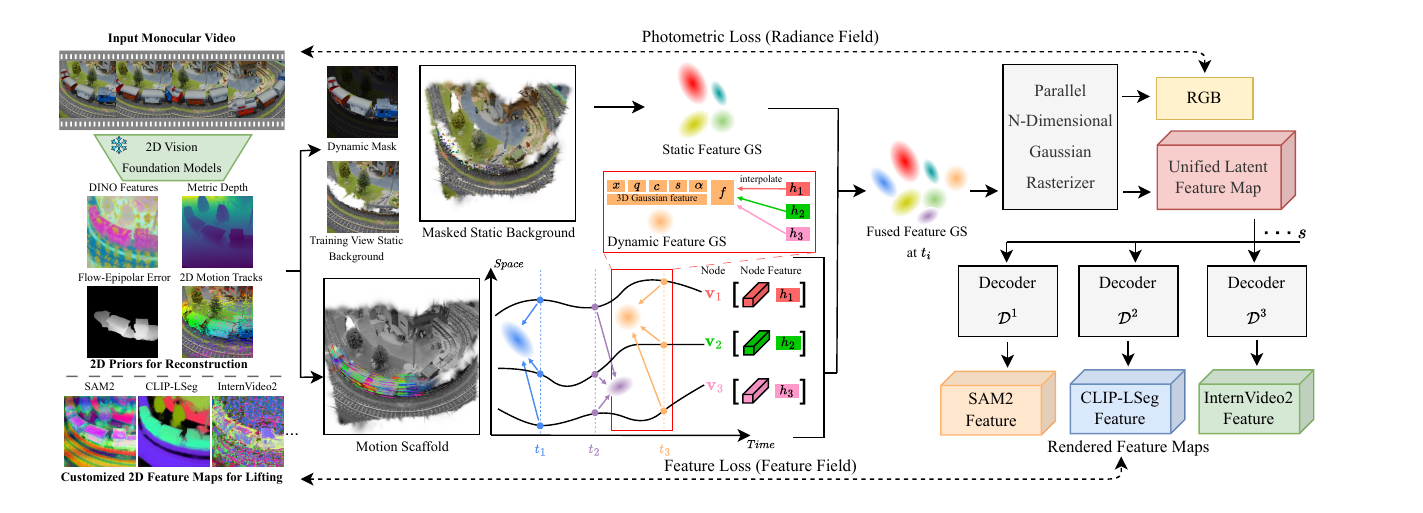}
   \caption{\textbf{Method overview.}
   Given an input monocular video, we infer 2D priors to segment static background (represented by static 3D Gaussians augmented with latent features) and dynamic foreground (represented by dynamic 3D Gaussians guided by Motion Scaffolds~\cite{lei2024mosca}, a set of nodes $\{\mathbf{v}_{i}\}$ encoding 3D motion trajectories and latent features $h_i$). Dynamic Gaussian features and motions are computed via interpolation from their $K$-nearest scaffold nodes. At each timestep, dynamic Gaussians are warped and fused with static Gaussians. A parallel rasterization~\cite{zhou2024feature} generates RGB images and a unified latent feature map, decoded into task-specific features—illustrated here by SAM2~\cite{ravi2024sam}, CLIP-LSeg~\cite{li2022language}, and InternVideo2~\cite{wang2024internvideo2} for representative 2D (novel view segmentation), 3D (scene editing), and 4D (spatiotemporal VQA) tasks. Our framework generalizes to any 2D vision foundation model and is trained end-to-end using input RGB frames and customized features from pretrained 2D models. At inference, rendered feature maps from arbitrary views and timesteps are directly fed into task-specific decoders, seamlessly supporting user prompts and LLM interactions to form a unified 4D agentic AI system.
   }
   \label{fig:pipeline}
   \vspace{-1.2mm}
\end{figure*}
\section{Related Works}
\label{sec:related}

\subsection{4D Representations for Reconstruction and Generation}
Various 4D representations are designed to support a wide variety of 4D tasks. Among them, 4D reconstruction is the most widely studied topic that drives the field. Early methods~\cite{gafni2021dynamic,pumarola2021d,peng2021animatable,park2021hypernerf,fridovich2023k,cao2023hexplane,kwon2021neural,noguchi2021narf,peng2021neuralbody,su2021anerf,weng2022humannerf,gafni2021nerface,li2021neural,du2021neural} implement variants of Neural Radiance Fields~\cite{nerf} to represent a dynamic 4D scene through implicit or hybrid representations. More recently, 3D Gaussian~\cite{kerbl20233d} has prevailed as a popular explicit 3D scene representation~\cite{chung2023luciddreamer, zhou2024dreamscene360, yu2024wonderworld} and various variants~\cite{luiten2023dynamic,wu20234d,duan20244d,yang2023real,yang2024deformable,huang2024sc,liang2023gaufre,wang2024shape,lei2024mosca,mihajlovic2025splatfields,stearns2024dynamic,ji2024segment} are similarly proposed to represent dynamic 4D scenes. 
Additionally, the field of 4D Generation~\cite{yin20234dgen,ren2023dreamgaussian4d,ren2024l4gm,ling2023align,xu2024comp4d} has witnessed great advancements in representations that are more suitable for generative settings.
In summary, most 4D representations can be categorized into deformation-based approaches~\cite{luiten2023dynamic,tretschk2021non,pumarola2020dnerf,park2021nerfies,park2021hypernerf,wu2022d,sun20243dgstream,li2024spacetime}, temporal extended approaches~\cite{yang2023real,cao2023hexplane,kplanes,duan20244d,zhang2024togs},  or an ensemble of multiple time-varying representations~\cite{xian2021space,li2020nsff,gao2021dynamic,li20244k4dgen,ren2024l4gm}.
In our work, we extend the direction of dynamic 3D Gaussian Splatting to a versatile Gaussian feature field that simultaneously embraces multiple features for various vision tasks beyond 4D reconstruction and generation. To the best of our knowledge, this direction has not been widely explored.


\subsection{4D Reconstruction from Monocular Video}
In the context of 4D reconstruction tasks, most methods~\cite{niemeyer2019occupancy,lin2024gaussian} focus on reconstructing dynamic 4D scenes from multiple well-calibrated videos. In contrast, another popular direction is to synthesize generic 4D scenes from monocular videos~\cite{xian2021space,tretschk2021non,pumarola2020dnerf,li2020nsff,park2021nerfies,gao2021dynamic,park2021hypernerf,wu2022d}. 
For unposed casual videos, non-rigid structure-from-motion workflows~\cite{blanz1999morphable,ramakrishna2012reconstructing,bogo2016keep,kar2015category,cmrKanazawa18,kulkarni2020acsm,yang2021lasr,yang2021banmo,curless1996volumetric,newcombe2015dynamicfusion,li2008global,gao2019surfelwarp,bozic2020neuraltracking,teed2021droid,zhao2022particlesfm} are adopted for estimating camera poses for each frame for further processing. Shape-of-motion~\cite{wang2024shape} and MoSca~\cite{lei2024mosca} develop end-to-end workflows that reconstruct 4D scenes from casual captured videos with the help of 3D Gaussians~\cite{kerbl20233d,lu2024scaffold,luiten2023dynamic}. Another series of work~\cite{yin20234dgen,li20244k4dgen,ren2023dreamgaussian4d,ren2024l4gm} assumes the camera is fixed for the video and reconstructs 3D in the camera coordinate space. More recently, the success of DUSt3R~\cite{wang2024dust3r}, a transformer-based novel paradigm for 3D reconstruction of arbitrary image collections, has been adapted for dynamic scenes in MonST3R~\cite{zhang2024monst3r}. 
In comparison, we aim to perform monocular 4D reconstruction and feature lifting simultaneously, which cannot be achieved in previous approaches.

\subsection{Feature Field Distillation}

Research in novel view synthesis and feature field representation has been extensively developed within the NeRF framework~\cite{mildenhall2021nerf} and 3D Gaussians~\cite{kerbl20233d}. Seminal works such as Semantic NeRF \cite{semantic_nerf} and Panoptic Lifting \cite{panoptic_lifting} have integrated semantic information from segmentation networks into 3D spaces, revealing that combining noisy or inconsistent 2D labels within a 3D context can produce clear and accurate 3D segmentations. Building on this concept, methods such as those in \cite{ren-cvpr2022-nvos} have shown that minimal input, like basic foreground-background masks, can be effective for object segmentation in 3D. Moving beyond label estimation to optimize NeRF, approaches like Distilled Feature Fields \cite{kobayashi2022decomposing}, NeRF-SOS \cite{fan2022nerf}, LERF \cite{kerr2023lerf}, and Neural Feature Fusion Fields \cite{tschernezki22neural} have embedded pixel-aligned feature vectors from tools such as LSeg or DINO \cite{dino} into NeRF structures. More recently, numerous works~\cite{yu2024language,shi2024language,zhou2024feature,qin2024langsplat,liu2024splatraj,zuo2024fmgs,lee2024rethinking} adopt similar strategies to distill information from well-trained 2D models to 3D Gaussians.
However, feature lifting into 4D fields is not properly solved yet. In this work, we make the first attempt to deliver a versatile 4D Gaussian feature field framework that embraces multiple task features simultaneously.

\section{Method}

Given a monocular RGB video $\mathcal{I} = {I_1, \ldots, I_t}$, we reconstruct the underlying 4D scene as a set of dynamic 3D Gaussians (\cref{sec:method_prelim}), each augmented with a unified latent feature embedding that distills informative features from various 2D foundation models for downstream tasks (\cref{sec:method_unified_latent}). To address the challenge of handling high-dimensional features across a large number of Gaussians and timesteps, we adopt a compact feature representation (\cref{sec:method_compact}).

\begin{figure}[t]
  \centering
   \includegraphics[width=1.0\linewidth]{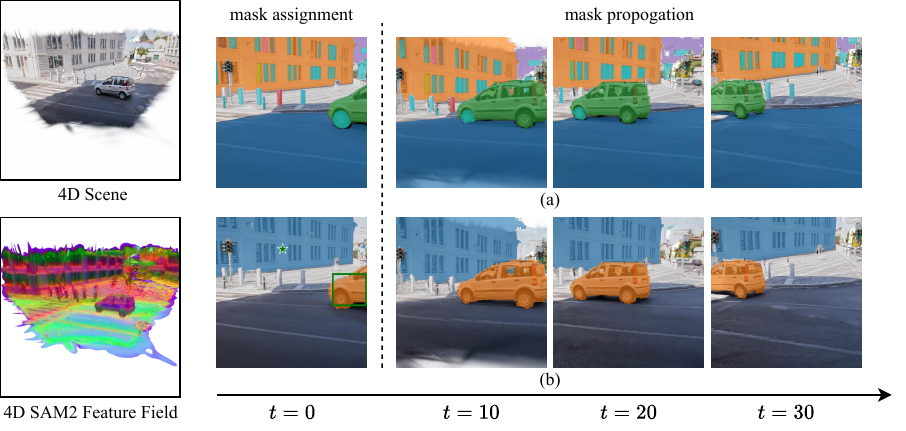}
   \caption{\textbf{Segment Anything in Dynamic 4D Scenes with SAM2 Feature Field.} For any rendered novel view video, we support: (a) Promptless segmentation (segment everything): when no user prompt is provided, segmentation masks are automatically assigned at the first frame ($t = 0$) and then propagated across all frames. (b) Promptable segmentation (segment anything): the user can segment any object—static or dynamic—at any timestep using a point or box prompt, and the corresponding mask is robustly tracked and propagated through subsequent frames.}
   \label{fig:sam2}
   \vspace{-4mm}
\end{figure}

\subsection{Preliminaries: Dynamic 3D Gaussian Splatting}\label{sec:method_prelim}

Following state-of-the-art monocular video-based dynamic 4D scene reconstruction approaches~\cite{wang2024shape, lei2024mosca, stearns2024dynamic}, we represent the scene with dynamic 3D Gaussians~\cite{luiten2023dynamic}, namely a set of persistent 3D Gaussians~\cite{kerbl20233d} that deform over time $t$. 
Specifically, we base our approach on MoSca~\cite{lei2024mosca}, which can reconstruct the dynamic 4D scene from monocular casual video. 
Like other methods, MoSca addresses the challenge from single-view partial observation by leveraging priors from 2D foundation models~\cite{unidepth,zoedepth,teed2020raft,cotracker,harley2022particle} and by regularizing Gaussian motion trajectories. 
At the core of MoSca's method is a structured graph named 4D Motion Scaffold $(\mathcal{V}, \mathcal{E})$ that drives the deformation of individual 3D Gaussians $\mathcal{G} = \{G_j\}_{j=1}^n$ (see~\cref{fig:pipeline}). Each node $\mathbf{v}^{(i)} \in \mathcal{V}$ describes a 3D motion trajectory $[\mathbf Q^{(i)}_1,\ldots,\mathbf  Q^{(i)}_t], \mathbf Q=[\mathbf R, \mathbf t]\in SE(3)$ with a control radius $r^{(i)}$. 
Edges $\mathcal{E}$ describe a $k$-nearest neighbor graph over the motion trajectories. 
Given any 3D Gaussian located at $\mathbf{x}$ at time $\tau$, 
to compute its deformation to time $\tau'$, 
we first find the trajectory node $\mathbf{v}^{(i^{*})}$ with the closest position $\mathbf p_{\tau}^{(i)}$ at time $\tau$, 
i.e., $i^* = \arg\min_{i} ||\mathbf p_{\tau}^{(i)} - \mathbf x||$, 
then interpolate the deformation from $K$ nearest trajectory nodes $\{\mathbf{v}^{(i)}\}_{i \in \mathcal{E}{(i^{*})}}$, with weights $\{w_i\}$ computed from the Gaussian-to-trajectory distance and the node control radius. 
In other words, the dense, per-Gaussian motion trajectories are interpolations of a much smaller ($100\times$) set of trajectory nodes. This design leverages the low-rank nature of the underlying scene motion and effectively regularizes Gaussian motions under limited supervision. 
We follow \cite{lei2024mosca}'s protocols to initialize and train 4D Motion Scaffolds and dynamic Gaussians that represent the dynamic foreground elements, as well as another set of static Gaussians that model the static background. Please refer to our supplementary material for full details.

\subsection{Unified Latent Feature Distillation} \label{sec:method_unified_latent}
We aim to go beyond appearance reconstruction by building a unified, versatile, and dynamic 4D feature field representation, capable of supporting diverse downstream visual tasks $\mathcal{T} = \{T^{1}, \ldots, T^{S}\}$ spanning 2D, 3D, and 4D, such as novel-view segmentation (2D), scene editing (3D), and scene-level spatiotemporal VQA (4D). A straightforward way to construct this 4D feature field is to extend existing 3D feature field frameworks~\cite{zhou2024feature} by replacing their 3D reconstruction modules with dynamic 4D reconstruction. However, this approach addresses each downstream task $T^s$ independently: input frames $\mathcal{I} = \{I_1, \ldots, I_t\}$ are encoded into task-specific 2D feature maps $\mathcal{F}^{s} = \{F^s_1, \ldots, F^s_{t}\}$ using dedicated vision foundation model encoders $E^s$, leading to separate feature fields per task. Consequently, performing three tasks would require three separate reconstructions, making this strategy inefficient and redundant.

To overcome this limitation, we propose distilling a unified 4D feature field to coherently fuse diverse 2D vision foundation model features across views and timesteps, providing consistent feature access in 4D (3D space + time) for various downstream tasks. Specifically, as shown in~\cref{fig:pipeline}, we utilize the parallel N-dimensional Gaussian rasterizer~\cite{zhou2024feature} to simultaneously render RGB images along with a unified latent feature $\mathcal{F}$, which is jointly learned with a set of lightweight decoders $\{\mathcal{D}^1, \ldots, \mathcal{D}^{S}\}$. Each decoder maps the shared latent feature $\mathcal{F} \in \mathbb{R}^{D}$ into task-specific target features $\mathcal{F}^{s} \in \mathbb{R}^{D_s}$, where $D$ and $D_s$ denote the dimensions of the unified latent and task-specific features, respectively, with $D \ll D_s$. Compared to separately optimizing high-dimensional features for each task, our unified latent representation significantly reduces computational overhead during rasterization. During the optimization process, we attach the feature vector $f_j \in \mathbb{R}^{D}$ to each 3D Gaussian $G_j \in \mathcal{G}$, warp $G_j$ to the target timestep $\tau$ following the process introduced in~\cite{lei2024mosca}, and rasterize $f_j$ the same way we rasterize Gaussian color $c_j$ as~\cite{zhou2024feature}. Conceptually, the RGB and feature reconstruction from viewpoint $v$ at timestep $\tau$ are computed as:
\begin{equation}
\begin{aligned}
\hat{I}_{\tau}^{v} & = \operatorname{Rasterize}(v, \{\operatorname{warp}(G_j, \tau), c_j)\}_{G_j \in \mathcal{G}}) \\
\hat{F}_{\tau}^{v} & = \operatorname{Rasterize}(v, \{\operatorname{warp}(G_j, \tau), f_j)\}_{G_j \in \mathcal{G}})
\end{aligned}
\end{equation}
The resulting feature map $\hat{F}_{\tau}$ (with viewpoint v omitted for simplicity) is decoded by $\mathcal{D}^s$ into a task-specific feature map $\hat{F}^{s}_{\tau}$, supervised by the ground-truth feature map obtained from a customized vision model encoder $E^{s}$ (e.g., SAM2). Concretely, we optimize the following feature loss $L_{\text{feat}}$ to jointly learn the feature field, alongside the photometric loss from~\cite{lei2024mosca} used for learning the radiance field.
\begin{align}
L_{\text{feat}} = \sum_{s=1}^S \operatorname{MSE}(\hat{F^s_{\tau}}, F^s_{\tau}), \\
\hat{F^{s}_{\tau}} = \mathcal{D}^s(\hat{F_{\tau}}), \quad \quad F^s_{\tau} = E^s(I_{\tau}).
\end{align}
Note that the feature loss and photometric loss are independent, and introducing the feature field does not degrade the quality of the radiance field reconstruction (see~\cref{tab:nvidia}).

\begin{figure}[t]
\begin{center}
 \includegraphics[width=\linewidth]{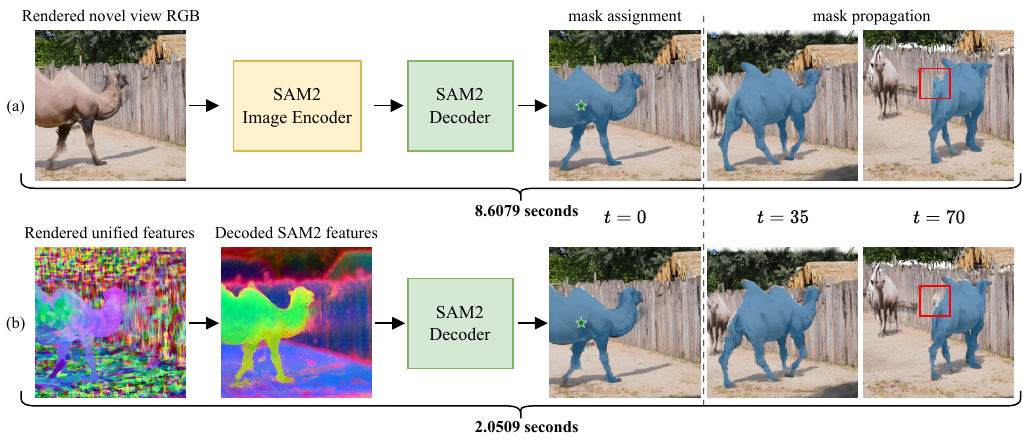}
  \caption{\textbf{Baseline Comparison on SAM2 Inference.} We compare segmentation quality and inference speed between (a) the naive RGB-based approach and (b) our feature-based method. Ours achieves comparable segmentation, accurately tracking the object over time, and avoids RGB artifacts (red box region at $t=70$), while reducing inference time to about 4$\times$ speed-up.}
  \label{fig:sam2_compare}
  \vspace{-5.5mm}
\end{center}
\end{figure}

\subsection{Scaffold-Based Compact Feature} \label{sec:method_compact}
While our unified feature field effectively reduces feature dimensionality, optimizing a feature vector for every Gaussian remains computationally expensive. However, semantic features tend to be smooth in 3D and exhibit strong local correlations, similar to Gaussian motion trajectories. To leverage this correlation, we propose representing per-Gaussian features $\{f_j\}$ as linear combinations of a smaller ($100\times$ fewer) and more compact set of base features $\{h_i\}$, conveniently attached to nodes $\{\mathbf{v}^{(i)}\}$ of the 4D Motion Scaffold (see~\cref{fig:pipeline}). Recall that we compute per-Gaussian deformations by interpolating from their $K$-nearest trajectory nodes $\{\mathbf{v}^{(i)}\}_{i \in \mathcal{E}(i^*)}$ with interpolation weights $\{w_i\}$. We reuse these same weights to obtain the per-Gaussian unified features as:
\begin{equation}
    f_j = \sum_{i \in \mathcal{E}(i^*)} w_i h_i.
\end{equation}
This compact feature representation significantly reduces the number of parameters required for optimization and provides structural regularization, encouraging smoother learned features.

\subsection{Interaction with AI Agent via Feature Fields}

Agentic AI typically requires cross-modal interactions, notably between language and vision. In this work, we aim to build an AI agent that 
supports direct language interactions with our dynamic 4D scene representation. Such interaction requires a shared feature space between text and Gaussian features. 
However, text features are usually high-dimensional. For instance, language-guided scene editing requires CLIP features of 512 dimensions. Directly assigning these high-dimensional features to each Gaussian is computationally expensive. Feature 3DGS~\cite{zhou2024feature} addresses this by assigning each Gaussian a full 512-dimensional CLIP feature via a parallel N-dimensional Gaussian rasterizer, enabling direct language interaction but resulting in slow rendering speeds and excessive memory usage. Although Feature 3DGS proposes a CNN-based acceleration, it operates only on 2D rendered maps, failing to directly resolve the feature mismatch in 3D. In contrast, we optimize a compact, lower-dimensional Gaussian feature ($D=32$), train an MLP-based decoder on rendered 2D features, and apply it directly to 3D Gaussian features during inference. Given the intrinsic flexibility of MLPs, our approach efficiently bridges language features and our compact feature fields, enabling direct interaction between LLMs and our 4D scene through language. Furthermore, by leveraging InternVideo2 features, which can be rendered and decoded from any viewpoint in 3D space over time, we naturally lift the video chatbot LLM from 2D to 4D, enabling free-form language interaction with the AI agent within 4D space.

Beyond direct language interaction, LLM agents can operate in the loop for manipulation tasks by interpreting user prompts, automatically optimizing hyperparameters, and iteratively refining results for downstream applications. For example, given a scene-editing prompt like “Delete the dog,” the agent parses the operation command (“delete”) and the target object (“dog”), then generates an editing configuration with relevant hyperparameters, such as the softmax threshold for matching Gaussians to the target object. It tests various thresholds, evaluates the quality of rendered image samples, and selects the optimal configuration. This is then consistently applied across the entire 4D scene, enabling efficient, intelligent editing with minimal user input.

This perception-reasoning-action loop empowers the LLM-driven 4D agent to interpret, execute, and refine complex scene manipulations, making it a powerful tool for adaptive 4D scene editing and interaction. The system is particularly valuable for applications like interactive VR content creation and editing, where dynamic scene understanding and precise, context-aware modifications are essential.


\section{Experiments}
In this section, we present experimental results and analyses of the capabilities of our proposed Feature4X framework. Based on the type of desired output, we categorize tasks into 2D (segmentation; see~\cref{sec:sam2,sec:lseg}), 3D (scene editing; see~\cref{sec:editing}), and 4D (spatiotemporal VQA; see~\cref{sec:vqa}).

\begin{figure}[t]
\begin{center}
 \includegraphics[width=\linewidth]{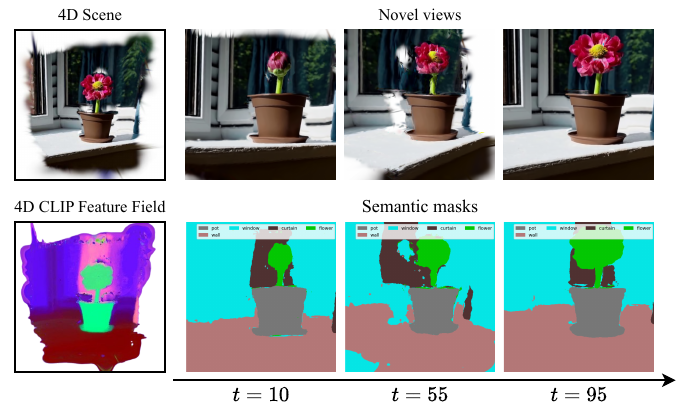}
 \vspace{-7mm}      
     \caption{\textbf{Semantic 4D Scene Understanding with CLIP Feature Field.} By lifting CLIP-LSeg~\cite{li2022language} features into a 4D feature field, we enable pixel-level semantic segmentation from any view at any timestep. This allows robust 4D scene understanding, even as object appearances change over time—for example, accurately identifying a blooming flower from bud to full bloom across views.}
  \label{fig:results2_sam_segementation_fea_img}
  \vspace{-4mm}
\end{center}
\end{figure}

\begin{table}[t]
\begin{center}
\resizebox{1.0\linewidth}{!}{
\begin{tabular}{lcccc} \toprule
Method & PSNR$\uparrow$  & mIoU$\uparrow$ & accuracy$\uparrow$ & Size (MB)$\downarrow$    \\
\midrule
 MoSca & 25.166 &   -  & -  &   \textbf{67.726}     \\
 MoSca + Feature 3DGS & 25.191 & 0.506 & \textbf{0.881} & 593.907   \\
 Ours (single CLIP head) & 25.186 & \textbf{0.510} & 0.880 &95.294       \\
 Ours (full model)                                      & \textbf{25.197} &0.503 & 0.876 & 95.457        \\
\bottomrule
\end{tabular}
}
\vspace{-5.5mm}
\end{center}
\caption{\textbf{Semantic segmentation on the Nvidia dataset~\cite{liu2023robust}.} Our method achieves comparable radiance reconstruction (PSNR) and segmentation performance, while significantly reducing memory usage compared to the baselines.}
\label{tab:nvidia}
\vspace{-5mm}
\end{table}

\subsection{Scene Interaction with Segment Anything}
\label{sec:sam2}

\begin{figure*}[t]
  \centering
   \includegraphics[width=\linewidth]{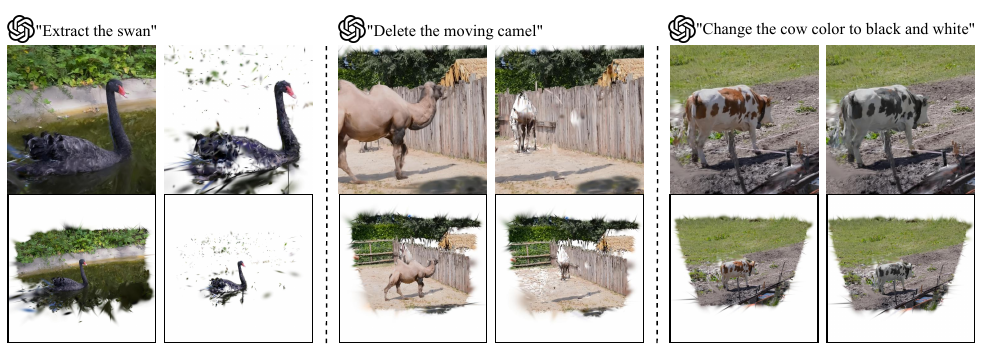}
   \caption{\textbf{Scene Editing with AI Agent.} Given user prompts, our GPT-powered agent interprets editing intent and autonomously performs scene edits via our 4D CLIP feature field. Examples include both geometric (e.g., ``extract'' and ``delete'') and appearance (e.g., ``change color'') editing in 3D space. While results may not be perfect due to imperfect fine-grained feature alignment and non-optimal editing parameter tuning, the agent adaptively refines parameters and applies edits consistently across views and time—greatly reducing the need for manual tuning—and demonstrates robust, interactive 4D scene manipulation.}
   \label{fig:lang_edit}
   \vspace{-4mm}
\end{figure*}

Segment Anything Model 2 (SAM2)~\cite{ravi2024sam} is an advanced segmentation model for promptable visual segmentation across images and videos, supporting various input types such as points and bounding boxes for interactive and precise results. In our approach, we extract per-frame features using SAM2’s image encoder and lift them into a 4D feature field, enabling direct decoding from novel-view rendered feature maps. This allows segmentation that is inherently aware of both viewpoint changes and temporal dynamics, as shown in~\cref{fig:sam2}. Masks are seamlessly propagated through space and time, regardless of object motion or camera trajectory, supporting both automatic and user-guided segmentation with temporal coherence.

Notably, our approach bypasses the need to first render RGB videos and then apply the full SAM2 pipeline (see (a) in~\cref{fig:sam2_compare}). We observe that artifacts in novel view RGB rendering can mislead SAM2 during encoding, resulting in ambiguous and inaccurate segmentation despite smoother masks. In contrast, our method achieves faster and more robust segmentation in 4D scenes by operating directly in feature space.


\subsection{Scene Understanding with Semantics}
\label{sec:lseg}
To achieve pixel-level semantic segmentation across any novel view and timestep, we extend the capabilities of CLIP-LSeg~\cite{li2022language} by lifting its features into a 4D CLIP feature field. CLIP-LSeg is a language-driven semantic segmentation model that leverages the Contrastive Language-Image Pre-training (CLIP) framework to align textual descriptions with visual content, enabling zero-shot segmentation by interpreting descriptive input labels. By lifting these 2D features into a 4D representation, our approach captures both spatial and temporal information, facilitating consistent semantic understanding in dynamic scenes, even when objects undergo significant appearance changes over time. As depicted in~\cref{fig:results2_sam_segementation_fea_img}, our model accurately identifies a flower throughout its blooming process—from bud to full bloom—across various viewpoints and timesteps. This demonstrates the model’s proficiency in maintaining consistent semantic understanding despite substantial visual transformations in dynamic scenes. 

\cref{tab:nvidia} presents a quantitative comparison between our method and baselines on the Nvidia dataset~\cite{liu2023robust}. “MoSca~\cite{lei2024mosca}” is the baseline 4D reconstruction method, while “MoSca + Feature 3DGS~\cite{zhou2024feature}” is a naive, non-compact baseline using RGB + feature rasterization. Our models—using a 32-dimension latent feature and 512-dimension CLIP-LSeg~\cite{li2022language} as the target—maintain comparable PSNR to MoSca, showing no loss in reconstruction quality. The single CLIP head variant achieves the highest mIoU, demonstrating effective feature distillation. Although the naive baseline yields slightly higher accuracy, our models are about 6.2$\times$ more space-efficient, with the full model delivering competitive performance while remaining compact and general-purpose.

\subsection{Scene Editing with Language and AI Agent}
\label{sec:editing}

We employ an LLM agent (GPT-4o~\cite{achiam2023gpt}) to interpret natural language prompts, optimize editing parameters, execute precise queries, and iteratively refine results for intelligent 3D scene editing. The agent begins by parsing the user’s high-level prompt to identify the editing action—such as ``extract,'' ``delete,'' or ``change color''—and the target object (e.g., ``cow''), as shown in~\cref{fig:lang_edit}. Based on the task, it generates candidate configurations with varying parameters relevant to the prompt. Specifically, we first compute the probability of each 3D Gaussian corresponding to a language description $l$ of the target object as:
\begin{equation}
\mathbf{p}(l \mid j)=\frac{e^s}{\sum_{s_i \in \mathcal{L}} e^{s_i}}, \quad s = \frac{q(l) \cdot f_j}{\|q(l)\|\|f_j\|},
\end{equation}
where $q(l)$ is the text feature, $f_j$ is the CLIP feature of Gaussian $j$, and $\mathcal{L}$ is the label set of all possible scene objects. The agent filters Gaussians using various thresholds based on the softmax probability $\mathbf{p}(l \mid j)$, applies the edit configuration, and evaluates the results via rendered image samples. It selects the parameter setting that best aligns with the intended edit and applies it consistently across all video frames, ensuring coherent editing throughout the 4D scene.

This GPT-powered 4D Agentic AI system interprets, executes, and refines complex scene manipulations, serving as an intelligent assistant for adaptive editing and interaction. By autonomously exploring, selecting, and tuning parameters—tasks typically handled by human editors—the agent enhances responsiveness and consistency. GPT acts as a self-refining decision-maker, illustrating the potential of agentic AI in creative and technical workflows where iterative optimization and goal-directed behavior significantly streamline and elevate editing tasks.

\subsection{Scene Reasoning with 4D Chatbot Agent}
\label{sec:vqa}


\begin{figure*}[t]
 \includegraphics[width=0.86\linewidth]
 {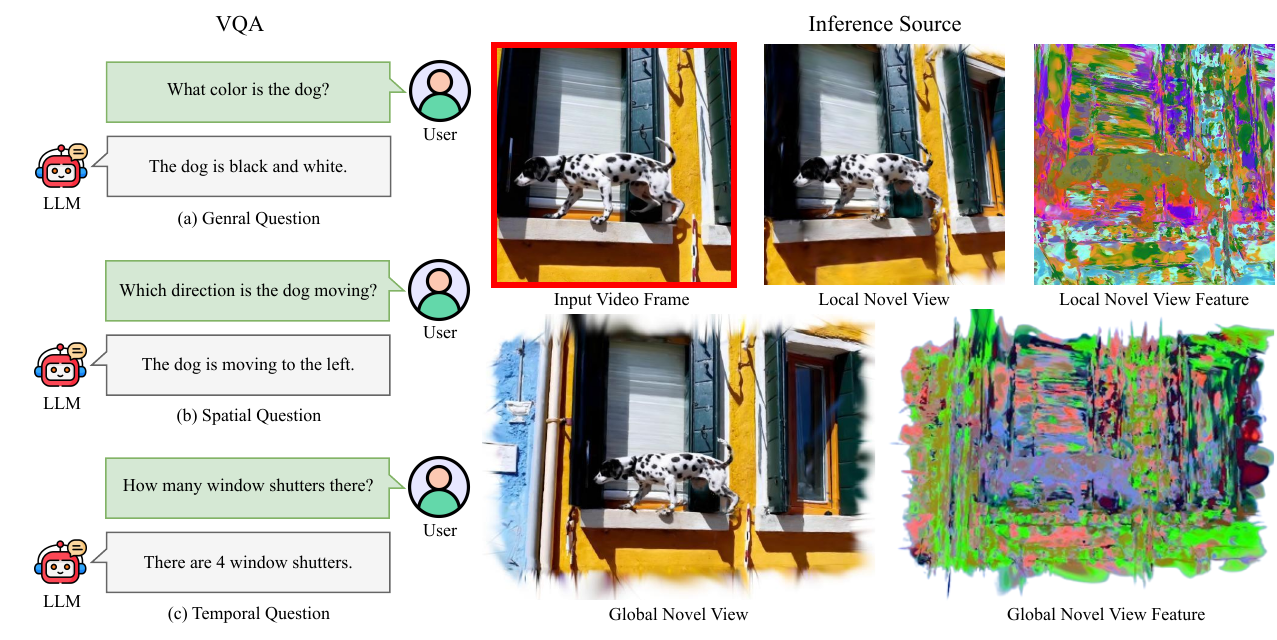}
 \centering
 \caption{\textbf{VQA with Chatbot Agent.} (Left) Our model supports free-form VQA across diverse question types—general, spatial, and temporal—by distilling InternVideo2~\cite{wang2024internvideo2} features. (Right) At each timestep, we reconstruct both a 4D radiance field and a 4D feature field, providing more inference sources beyond the input video frame—including local (moving camera) and global (zoomed-out) novel views and their corresponding feature maps—thereby supporting VQA in 4D and enhancing the model’s spatiotemporal reasoning capabilities.}
 \label{fig:vqa}
 \vspace{-3mm}
\end{figure*}

While large vision-language models (VLMs) have demonstrated impressive performance in visual question answering (VQA) tasks, their application has predominantly been confined to 2D visual modalities, such as images and videos. To our knowledge, we are the first to extend the capabilities of a VLM (Video-LLM) into 4D—encompassing three spatial dimensions plus time—by integrating VLM knowledge into our 4D feature field. This advancement enables a more comprehensive understanding and interaction with dynamic 4D scenes, capturing both spatial and temporal nuances beyond the static 2D perspective.

We utilize InternVideo2~\cite{wang2024internvideo2,li2023videochat}–a state-of-the-art video foundation model–to extract comprehensive video features. Specifically, we sample the input video into several clips, each comprising 8 frames sampled at equal intervals to ensure enough GPU memory available during the whole process. These clips are processed through InternVideo2’s vision transformer encoder to obtain segment-level features and then aggregated together. Additionally, for each segment, InternVideo2 generates a class token representing high-level semantic information. We compute the average of these class tokens to form a global video representation. This class feature is then concatenated with the aggregated segment features, effectively integrating temporal information which is necessary for dynamic scene reasoning.

In our system, free-form visual question answering (VQA) is supported through a language model~\cite{li2023videochat}, enabling users to interact with a chatbot agent using natural language while exploring 4D dynamic scenes, as illustrated in~\cref{fig:vqa}. Unlike prior approaches that rely solely on 2D monocular video—limited to a single view at a single time step—our method leverages the reconstructed 4D radiance and feature fields to support richer inference sources, including both local novel views (via free camera movement) and global novel views (via zoom-out for broader context). In the example shown in~\cref{fig:vqa}, we observe that InternVideo2, when operating on the input monocular video, fails to correctly answer spatial and temporal questions—even when the answers are obvious to humans—while the same model, using our global novel view feature, succeeds (see~\cref{tab:4D Q&A}). This is because monocular video lacks the spatial coverage and contextual cues necessary for spatiotemporal reasoning. In contrast, our framework allows rendering of novel views and their corresponding feature maps at arbitrary time steps, injecting explicit signals that enhance reasoning. To evaluate this more systematically, we construct 400 objective questions across 50 scenes in the DAVIS dataset~\cite{perazzi2016benchmark}, covering spatial only (e.g., “Is the person facing left?”), temporal only (e.g., “How many times does the dancer spin?”), and spatiotemporal (e.g., “Which direction is the duck moving?”) categories. As shown in~\cref{tab:vqa_davis}, our method improves VQA accuracy by enriching scene understanding through 4D feature field reconstruction, while also accelerating inference by eliminating the need for video encoding.

\begin{table}[t]
\renewcommand{\arraystretch}{0.8} 
\begin{center}
\resizebox{1.0\linewidth}{!}{
\begin{tabular}{l|ccc|c} \toprule
Inference Source & General & Spatial & Temporal & Time (s) \\
\midrule
 Input Video View & \checkmark & $\times$ & $\times$ & 33.65 \\
 Local Novel View & \checkmark & \checkmark & $\times$ & 33.86 \\
 Local Novel Feature & \checkmark & $\times$ & \checkmark & 11.99 \\
 Global Novel View & \checkmark & $\times$ & \checkmark & 33.37 \\
 Global Novel Feature & \checkmark & \checkmark & \checkmark & 12.24 \\
\bottomrule
\end{tabular}}
\vspace{-5mm}
\end{center}
\caption{\textbf{VQA performance across different inference sources with the scene shown in~\cref{fig:vqa}.} Feature-based inference supports all question types with lower latency, while view-based methods are limited in spatial and temporal reasoning.}
\label{tab:4D Q&A}
\vspace{-3mm}
\end{table}

\begin{table}[t]
\begin{center}
\resizebox{1.0\linewidth}{!}{
\begin{tabular}{@{}lcccc@{}} 
\toprule
Inference Source & Spatial Acc. & Temporal Acc. & Overall Acc. & Time (s)\\
\midrule
 Input Video View & 48.50 & 49.84 & 49.06 & 10.02\\
 Local Novel View & 49.75 & 51.13 & 50.31 & 9.78\\
 Local Novel Feature & 54.50 & 54.37 & 56.29 & \textbf{2.81}\\
 Global Novel View & 47.00 & 49.19 & 47.48 & 10.06\\
 Global Novel Feature & \textbf{58.75} & \textbf{58.25} & \textbf{61.32} & 3.42\\
\bottomrule
\end{tabular}
}\vspace{-5mm}
\end{center}
\caption{\textbf{Spatiotemporal VQA on DAVIS dataset~\cite{perazzi2016benchmark}.} Compared to 2D video inference, our 4D feature space inference (Global Novel Feature) enhances the Video-LLM’s spatiotemporal reasoning while achieving approximately 3$\times$ faster inference.}
\label{tab:vqa_davis}
\vspace{-5mm}
\end{table}

\section{Conclusion}

We have introduced Feature4X, a general and scalable framework that bridges casually captured monocular videos to interactive 4D agentic AI systems. By distilling diverse 2D vision foundation model features into a unified, dynamic 4D Gaussian feature field, our method supports a wide range of tasks across 2D (segmentation), 3D (scene editing), and 4D (spatiotemporal VQA). Through a compact and versatile representation, Feature4X achieves efficient training and inference while maintaining high-quality appearance reconstruction and robust semantic understanding. Furthermore, by integrating LLMs for language-based interaction and autonomous task refinement, our system elevates traditional perception tasks into a perception-reasoning-action loop, enabling intelligent and adaptive manipulation of dynamic 4D scenes. We hope Feature4X opens new avenues in 4D vision and agentic AI research by facilitating immersive, multimodal interaction with dynamic visual content.

\vspace{-1.7mm}
\paragraph{Acknowledgement} 
We especially thank Jiahui Lei for valuable guidance on MoSca implementation and insightful suggestions on experiments. This project was supported by LUCI program under the Basic Research Office and partially supported by ARL grants W911NF20-2-0158 and W911NF-21-2-0104 under the cooperative A2I2 program. Z.W. is supported by ARL and LUCI program, and an Army Young Investigator Award. L.G. is supported by a Vannevar Bush Faculty Fellowship. A.K. is supported by a DARPA Young Faculty Award, NSF CAREER Award IIS-2046737, and Army Young Investigator Award.
\label{sec:acknowledgement}

\clearpage
{
    \small
    \bibliographystyle{ieeenat_fullname}
    \bibliography{main}
}

\clearpage
\setcounter{section}{0}
\setcounter{figure}{0}
\setcounter{table}{0}
\maketitlesupplementary

\renewcommand\thesection{\Alph{section}} 
\renewcommand\thesubsection{\thesection.\arabic{subsection}} 
\renewcommand\thefigure{\Alph{figure}} 
\renewcommand\thetable{\Alph{table}} 

\crefname{section}{Sec.}{Secs.}
\Crefname{section}{Section}{Sections}
\Crefname{table}{Table}{Tables}
\crefname{table}{Tab.}{Tabs.}

\newcommand{\tabnohref}[1]{Tab.~{\color{red}#1}} 
\newcommand{\fignohref}[1]{Fig.~{\color{red}#1}} 
\newcommand{\secnohref}[1]{Sec.~{\color{red}#1}} 
\newcommand{\cnohref}[1]{[{\color{green}#1}]} 
\newcommand{\linenohref}[1]{Line~{\color{red}#1}}

\noindent This supplement is organized as follows:
\begin{itemize}[itemsep=0em]
    \item Section~\ref{sec:Details of MoSca} contains more details of 4D reconstruction;
    \item Section~\ref{sec:Details of Architectures} contains implementation details;
    \item Section~\ref{sec: Foundation Models Features} contains the details of foundation models' features extraction and downstream task inference;
    \item Section~\ref{sec:Editing} contains the algorithmic details of language-guided 4D editing with LLM;
    \item Section~\ref{sec:Baseline Comparisons} contains more baseline comparisons;
    \item Section~\ref{sec:Ablation Studies} contains ablation studies of our method.
    
\end{itemize}

\section{Details of 4D Reconstruction}
\label{sec:Details of MoSca}


Our video-to-4D feature field reconstruction pipeline is based upon MoSca~\cite{lei2024mosca}, a state-of-the-art monocular video-based dynamic 3D scene reconstruction method. We augment their representation with our proposed unified feature field and follow their training procedure with added feature rendering loss to the original optimization objective. Here we give a brief overview of the training pipeline. Please refer to MoSca~\cite{lei2024mosca} for more details.

\subsection{Dynamic Scene Representation}

Given a monocular input video, the underlying dynamic 3D scene is modeled as the composition of a static 3D background, represented with a set of static 3D Gaussians $\{\mathcal{G}_{\text{static}}\}$, and a dynamic 3D foreground, represented by a set of 3D Gaussians $\{\mathcal{G}\}$ that deform over time. 

The deformation of Gaussians $\{\mathcal{G}\}$ is modeled by a structured representation named 4D Motion Scaffold. It is a graph $(\mathcal{V}, \mathcal{E})$ where the nodes $\mathcal{V}$ are 3D motion trajectories $\mathbf{v}^{(i)} = [\mathbf Q^{(i)}_1,\ldots,\mathbf  Q^{(i)}_t], \mathbf Q=[\mathbf R, \mathbf t]\in SE(3)$. Intuitively, they describe the rigid transformations or 6DoF poses of points through time. They also each have a control radius attribute $r^{(i)}$, describing the range of influence.

Given two nodes, $\mathbf{v}^{(i)}, \mathbf{v}^{(j)}$, we define their distance as the maximum distance of their translation component $\mathbf{t}$ over all timesteps. Specifically, $D(i, j) = \max_{\tau} ||\mathbf{t}_{\tau}^{(i)} - \mathbf{t}_{\tau}^{(j)}||$. Intuitively, two nodes are close to each other only if they are close at all timesteps. Based on distance metric $D$, we construct a K-Nearest Neighbor (KNN) Graph $(\mathcal{V}, \mathcal{E})$, describing the mutual influence of nodes on each other. 

The set of dynamic 3D Gaussians $\mathcal{G}$ can be thought of as the union of 3D Gaussians from all timesteps. Each Gaussian $G \in \mathcal{G}$ is originally spawned at a certain source timestep $\tau$, with position $\mu$ and rotation $\mathbf{R}$. However, to render the scene at target timestep $\tau'$, we \textit{fuse} Gaussians from other source timesteps as well. This helps with the partiality of the single view observation at $\tau'$. To do so, we need to compute the motion or the deformation of Gaussian $G$ from $\tau$ to $\tau'$, which we achieve by querying the Motion Scaffold $(\mathcal{V}, \mathcal{E})$. 

From a high level, we find the node trajectories closest to $G$ and use the interpolation of their deformation (given by $\mathbf{Q}_{\tau'}\mathbf{Q}_{\tau}^{-1}$) as the deformation of $G$. For computation efficiency, in practice, we first identify the node trajectory $\mathbf{v}^{(i^{*})}$ closest to $g$ based on their positions at $\tau$. Specifically, $i^* = \arg\min_{i} ||\mathbf t_{\tau}^{(i)} - \mu||$. We then use $i^*$'s K-Nearest neighbors $\{\mathbf{v}^{(i)}\}_{i \in \mathcal{E}{(i^{*})}}$ as an approximation of $G$'s closest nodes. We compute the interpolation weights $\{w_i\}$ as
\begin{equation*}\label{eq:weights}
w_i = \operatorname{Normalize}_{i \in \mathcal{E}{(i^{*})}} \left(\exp (-\frac{||\mu - \mathbf{t}_{\tau}^{(i)}||^2_2}{2r^{(i)}}) + \Delta w_i\right),
\end{equation*}

Intuitively, nodes with closer spatial positions and larger control radius have more influence over $G$. $\Delta w_i$ is a learnable offset jointly optimized with other model parameters, allowing more flexibility.

The transformation of $G$ from time $\tau$ to $\tau'$ is computed as
$$\mathbf{T}_{\tau\to\tau'} = \operatorname{DQB}(\{w_i, \mathbf{Q}_{\tau'}\mathbf{Q}_{\tau}^{-1}\}) \in SE(3),$$
where $\operatorname{DQB}$ represents Dual Quaternion Blending that interpolates $SE(3)$ elements. We apply $\mathbf{T}_{\tau \to\tau'}$ to the position and rotation of $G$ when rendering it at timestep $\tau'$.

\subsection{Initialization from Lifted 2D Priors}

We rely on 2D priors from pretrained foundation models to initialize and constrain the system. The most important priors include 1) per-frame monocular metric depths $\mathcal{D}$ for rough geometry initialization; 2) long-term pixel trajectories $\mathcal{T}$, where each trajectory describes a point's pixel position at each frame, providing a rough motion initialization in 2D; and 3) per-frame epipolar error maps computed from dense optical flow, separating static background and dynamic foregrounds. 

We assume known camera intrinsic and poses for the following and refer the readers to \cite{lei2024mosca} for camera initialization and optimization in case the cameras are unknown. 

We first compute foreground-background masks from epipolar error maps $\mathcal{M}$. We initialize the static background 3D Gaussians by back-projecting points in the static background masks using depth estimations.

For the dynamic foreground, we lift 2D pixel trajectories to 3D space with monocular depth predictions and filter out dynamic ones based on epipolar error maps, then sample a subset of them based on trajectory distance metric $D$ as the initial Motion Scaffold nodes. As these trajectories only contain 3D positions, we initialize the rotational components $\mathbf{R}$ with identity matrices. We initialize the dynamic Gaussians by back-projecting points in the dynamic foreground masks.

\subsection{Optimization}

We first optimize the static background 3D Gaussians following the standard 3D Gaussian Splatting optimization procedures, using supervision from static regions of the input video frames. 

The dynamic foreground is optimized in two stages. The first geometric optimization stage focuses on the Motion Scaffolds and produces a reasonable deformation field, based on which the second photometric/feature stage spawns dynamic Gaussians and optimizes the final RGB and feature rendering objectives. During geometric optimization, the Scaffold trajectories are optimized subject to a set of physics-inspired losses: an as-rigid-as-possible loss that encourages nearby nodes to deform in locally-rigid ways, and smoothness losses on the velocity and acceleration of points to encourage smooth motions. During photometric optimization, we deform the dynamic Gaussians with Motion Scaffolds, combine them with the static Gaussians, rasterize them into RGB images and feature maps, and jointly optimize all parameters including Gaussian parameters and the Motion Scaffolds.




\section{Implementation Details}
\label{sec:Details of Architectures}
\paragraph{Feature Decoder Configuration}
While training our unified latent feature field, the rendered feature map from a camera view is passed through a decoder head to obtain the final feature map for each feature type. Our decoder heads are simple MLPs, which enlarge the input feature dimension by a factor of 2 at each layer. For example, our latent feature dimension is 32 but the CLIP feature dimension is 512, so the decoder simply maps 32 to 64, 64 to 128, and finally 128 to 512. Between each linear layer in the decoder, we use a ReLU activation function.


\paragraph{N-Dimensional Feature Map Rendering}
We utilize the parallel N-dimensional Gaussian rasterizer from Feature 3DGS~\cite{zhou2024feature}, which leverages a point-based $\alpha$-blending approach to rasterize feature maps. This method ensures that the rendered feature maps and RGB images are produced at the same resolution. Before calculating the loss, we align the size of the rendered feature map with the ground truth feature map. For SAM2~\cite{ravi2024sam} and CLIP-LSeg~\cite{li2022language} features, this is achieved using bilinear interpolation, while area resampling is applied to InternVideo~\cite{wang2024internvideo2} features due to the patchify mechanism used during feature extraction. Moreover, the parallel N-dimensional rasterizer is designed with adaptability, enabling flexible adjustments to the various dimensions of our unified latent feature field.


\paragraph{Training}
Our model is trained end-to-end to jointly reconstruct the versatile Gaussian feature field and the radiance field. We adopt the same training schedule as MoSca~\cite{lei2024mosca}: 4000 iterations for static Gaussian optimization and 5000 for dynamic one on the DAVIS dataset~\cite{perazzi2016benchmark}, and 8000 iterations for static and 5000 for dynamic Gaussian optimization on the NVIDIA dataset~\cite{yoon2020novel}.


\begin{figure*}[t]
  \centering
  \includegraphics{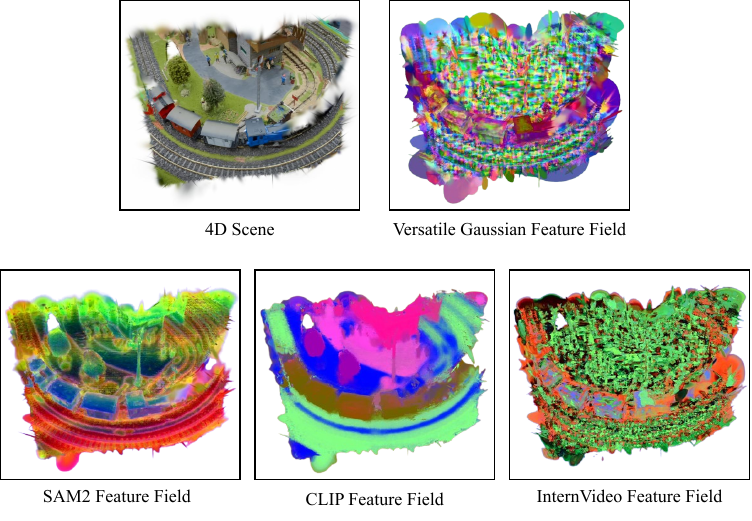}\vspace{-3mm}
  \caption{\textbf{Feature Field Visualizations}. We visualize our versatile Gaussian feature field along with its decoded SAM2, CLIP, and InternVideo feature fields using PCA.} 
  \label{fig:feature_visualization}
\end{figure*}

\section{Feature Extraction and Inference with Foundation Models}
\label{sec: Foundation Models Features}
\paragraph{SAM2 Feature} We extract per-frame ground truth features sequentially using the image encoder from SAM2~\cite{ravi2024sam}. Specifically, each frame is processed by a Hiera~\cite{ryali2023hiera} image encoder, which is pre-trained with MAE~\cite{he2022masked}, to produce a feature map with a resolution of $64 \times 64$ and a feature dimension of 256. Consequently, the ground truth features for the input video have a shape of $T \times 256 \times 64 \times 64$, where $T$ denotes the number of frames.

Our promptable / promptless segmentation results are obtained by feeding the rendered SAM2 feature map into the SAM2 decoding architecture, which includes a memory attention module, a prompt encoder, and a mask decoder. These components interact with a memory encoder and memory bank to retain and propagate segmentation information. For promptable segmentation, points, boxes, or masks are input into the prompt encoder to define the object's extent in a frame. For promptless segmentation, SAM2’s automatic mask generator produces segmentation masks for a frame, which are then input into the prompt encoder. Once segmentation is performed on any initial frame, SAM2's memory modules enable automatic tracking and propagation of masks across subsequent frames by conditioning each frame’s features on past predictions.

\paragraph{CLIP-LSeg Feature} For CLIP-LSeg, we utilize the CLIP ViT-L/16 image encoder to generate ground truth per-pixel feature maps and the ViT-L/16 text encoder for text feature extraction. This encoder can automatically resize any input image to generate a feature map with the longer side set to 480. We use square input images in our experiments so the CLIP-LSeg image encoder produces ground truth features with a resolution of $480 \times 480$ and a feature dimension of $512$. 

During inference for semantic segmentation, the rendered features with shape $(512, 480, 480)$ are reshaped into $(480 \times 480, 512)$, referred to as the image feature. Meanwhile, the text feature extracted from the CLIP text encoder has a shape of $(C, 512)$, where $C$ is the number of categories. A matrix multiplication is then performed between the image feature and the text feature to align pixel-level features with text queries. The resulting features are further processed using LSeg spatial regularization blocks to generate the semantic segmentation masks.

\paragraph{InternVideo Feature}
Given an input video, internvideo first resizes it to $224 \times 224$ resolution, then takes $14 \times 14$ as a patch, passing through a convolutional neural network to get the initial feature map with 1408 channels. Then, a pretrained class token input is concatenated to the initial feature map and passed through a transformer encoder to get the final feature map, which is in the shape of $(T \times 16 \times 16 + 1) \times 1408$, where $T$ is the number of frames. As the class token represents the whole video class information and is not dependent on individual pixels, we save and detach it when distilling the 4D feature field. 

During inference, we first concatenate the class token back into the novel-view rendered InternVideo feature to obtain the gathered feature. This gathered feature is then directly input into a Video-LLM~\cite{li2023videochat} to perform free-form visual question answering (VQA). Since the feature map can be rendered from any viewpoint of the 4D scene at high speed, our approach enables a seamless connection between the 4D scene and the AI agent (chatbot).

\paragraph{Feature Fields Visualization} As shown in~\cref{fig:feature_visualization}, we leverage Principal Component Analysis (PCA) from the scikit-learn library~\cite{pedregosa2011scikit} to visualize feature fields from global novel views. We configure the PCA to use 3 components corresponding to RGB channels and compute the PCA mean by sampling every third element along the $h \times w$ vectors. These vectors have feature dimensions of 32 (for unified latents), 256 (for SAM2), 512 (for CLIP), and 1408 (for InternVideo). The feature map is then transformed using the calculated PCA components and mean. This process involves centering the features using the PCA mean, followed by projection onto the PCA components. The transformed feature is subsequently normalized by removing outliers and scaling based on the minimum and maximum values, standardizing the feature values into a uniform range suitable for visualization.

\section{Details of LLM-powered 4D Editing}
\label{sec:Editing}

\begin{figure*}[t]
  \centering
  \includegraphics{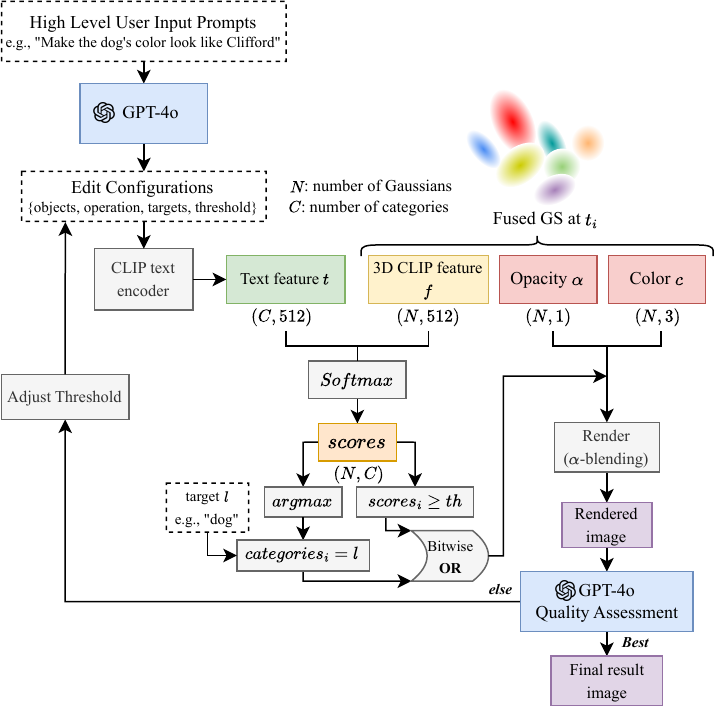}
  \caption{\textbf{Overview of the editing framework.} GPT-4o generates different editing configurations based on user prompts, selects target regions via hybrid filtering, evaluates their outputs, and selects the best configuration.} 
  \label{fig:edit_procedures}
\end{figure*}

An overview of the editing pipeline is illustrated in~\cref{fig:edit_procedures}.
It begins with user inputs, such as ``Make the dog's color look like Clifford," which are processed by a GPT-4o model to extract editable configurations (e.g., objects, operations, targets, and thresholds).
Each Gaussian $x_i$ is defined by $(f_i, \alpha_i, c_i)$, where $f_i \in \mathbb{R}^{512}$ is the semantic feature, $\alpha_i \in \mathbb{R}$ is the opacity, and $c_i \in \mathbb{R}^3$ is the color.
Guided by input specified object categories (e.g., ``dog", ``cow"), the CLIP ViT-B/32 text encoder encodes the text into features $\{t_1, \ldots, t_C\}$, where $t_i \in \mathbb{R}^{512}$ and $C$ is the number of categories. The inner product between text and semantic features, followed by a softmax, produces a semantic score matrix $scores \in \mathbb{R}^{N \times C}$. 

To select Gaussians that corresponds to the editing, we use a combination of two masking schemes derived from the semantic scores. The first one is binary thresholding. 
The query label $l \in \{1, 2, \ldots, C\}$ (or $l \subseteq \{1, 2, \ldots, C\}$ if it is a list of categories) determines the corresponding column of the score matrix $score_l = [s_{1l}, s_{2l}, \ldots, s_{Nl}]^{\top}$. Indices $i$ where $s_{il} \geq th$ are selected (set to 1), while the rest are excluded (set to 0). The selected indices define the target region, and unselected Gaussians are masked out.
The second masking scheme is determined by assigning each Gaussian to the category with the highest score using $argmax$ as shown in~\cref{fig:edit_procedures}, producing a category vector $categories = [c_1, c_2, \ldots, c_N]^{\top}$, where $c_i = \operatorname{argmax}\{s_{i1}, \ldots, s_{iC}\}$. Gaussians are selected if their category aligns with the query label $l$, and others are excluded.
Their resulting masks are combined using a bitwise OR operation. This method balances flexibility and precision, enabling more robust selection.

To optimize editing parameters, GPT-4o generates multiple candidate configurations with varying thresholds or transformations. Each candidate is used to execute an editing operation, and the resulting edited images are rendered. GPT-4o then evaluates these outputs and selects the best configuration, which is subsequently applied across the entire video to ensure consistent and high-quality results.

\begin{table*}[h]
  \begin{center}
  \setlength\tabcolsep{1pt}
    \resizebox{0.475\linewidth}{!}{
  \begin{tabular}{lccccccc} 
    \toprule
     \textbf{NVIDIA} & Exp1 & Exp2 & Exp3 & Mean$\uparrow$& Time (s)$\downarrow$\\
    \midrule
   RGB & 0.656 & 0.246 & 0.467 & 0.456 & 1.83\\
   Feature & \textbf{0.761} & \textbf{0.728} & \textbf{0.727} & \textbf{0.739} & \textbf{1.01}\\
    \bottomrule
  \end{tabular}} 
  \setlength\tabcolsep{1pt}
  \resizebox{0.45\linewidth}{!}{
  \begin{tabular}{lccccccccc} 
    \toprule
     \textbf{Nerfies} & Exp1 & Exp2 & Exp3 & Mean$\uparrow$& Time (s)$\downarrow$\\
    \midrule
  RGB     & 0.484 & 0.536          & 0.538          & 0.519          & 9.10\\
  Feature & \textbf{0.560}          & \textbf{0.662} & \textbf{0.561} & \textbf{0.594} & \textbf{3.10} \\
    \bottomrule
  \end{tabular}} 
  \caption{\textbf{SAM2 Quantitative Results (mIoU) on NVIDIA and Nerfies Datasets.}}
 \label{tab:sam2}
 \end{center}
\end{table*}

\begin{table*}[t]
\begin{center}
\resizebox{\linewidth}{!}{
\begin{tabular}{lccccccccc} \toprule
Scene & Method & PSNR$\uparrow$ & SSIM$\uparrow$ & LPIPS$\downarrow$          &accuracy$\uparrow$ & mIoU$\uparrow$ & Static Model Size (MB) & Dynamic Model Size (MB) & Size (MB)   \\
\midrule
Jumping & MoSca~\cite{lei2024mosca}                      &24.558     &0.792	&0.092	&- 		  &-       &29.08  &29.70    &58.78  \\
Jumping & MoSca + Feature 3DGS~\cite{zhou2024feature}    &24.516	    &0.793	&0.092	&0.840	&0.483	&271.30	&212.58   &483.88 \\
Jumping & Ours (single CLIP head)                        &24.633	    &0.795	&0.090	&0.836  &0.495	&44.15  &30.51    &74.66  \\
Jumping & Ours (full model)                                     &24.616	    &0.793	&0.090	&0.831	&0.483	&44.09  &30.84    &74.93  \\
\midrule
Skating & MoSca~\cite{lei2024mosca}                      &31.478     &0.926	&0.059	&- 		  &-       &32.49  &4.90     &37.39 \\
Skating & MoSca + Feature 3DGS~\cite{zhou2024feature}    &31.568	    &0.927	&0.059	&0.835	&0.446	&302.50 &35.40    &337.90 \\
Skating & Ours (single CLIP head)                        &31.572	    &0.927	&0.059	&0.838  &0.450	&49.32  &4.90     &54.22 \\
Skating & Ours (full model)                                          &31.666	    &0.926	&0.059	&0.819	&0.418	&47.52  &5.60     &53.12 \\
\midrule
Truck & MoSca~\cite{lei2024mosca}                      &26.688     &0.824	&0.115	&- 		  &-       &38.31  &11.84    &50.15 \\
Truck & MoSca + Feature 3DGS~\cite{zhou2024feature}    &26.619	    &0.824	&0.115	&0.973	&0.880	&353.97 &90.18    &444.15  \\
Truck & Ours (single CLIP head)                        &26.630	    &0.820	&0.122	&0.971  &0.878	&58.35  &12.27    &70.62 \\
Truck & Ours (full model)                                          &26.610	    &0.822	&0.117	&0.969	&0.868	&58.16  &13.51    &71.67 \\
\midrule
Umbrella & MoSca~\cite{lei2024mosca}                      &23.355      &0.706	&0.185	&- 		  &-       &71.29   &11.42    &82.71  \\
Umbrella & MoSca + Feature 3DGS~\cite{zhou2024feature}    &23.362	    &0.708	&0.176	&0.875	&0.556	&657.96  &75.12    &733.08    \\
Umbrella & Ours (single CLIP head)                        &23.433	    &0.707	&0.185	&0.869  &0.559	&107.87  &11.28    &119.15  \\
Umbrella & Ours (full model)                                          &23.392	    &0.708	&0.180	&0.880	&0.565	&107.58  &11.50    &119.08  \\
\midrule
Balloon1 & MoSca~\cite{lei2024mosca}                      &22.666      &0.760	&0.117	&- 		  &-      &56.88   &18.59    &75.47  \\
Balloon1 & MoSca + Feature 3DGS~\cite{zhou2024feature}    &22.687	    &0.762	&0.115	&0.901	&0.377	&525.09  &129.88   &654.97     \\
Balloon1 & Ours (single CLIP head)                        &22.668	    &0.760	&0.118	&0.905  &0.435	&85.95   &18.84    &104.79   \\
Balloon1 & Ours (full model)                                          &22.691	    &0.759	&0.117	&0.903	&0.446	&85.72   &19.90    &105.62   \\
\midrule
Balloon2 & MoSca~\cite{lei2024mosca}                      &26.827      &0.850	&0.082	&- 		  &-      &51.82   &15.22   &67.04  \\
Balloon2 & MoSca + Feature 3DGS~\cite{zhou2024feature}    &27.018	    &0.854	&0.080	&0.819	&0.350	&475.71  &108.96  &584.67      \\
Balloon2 & Ours (single CLIP head)                        &26.904	    &0.851	&0.081	&0.821  &0.321	&78.22   &15.18   &93.40    \\
Balloon2 & Ours (full model)                                          &26.871	    &0.853	&0.078	&0.813	&0.319	&77.77   &15.36   &93.13    \\
\midrule
Playground & MoSca~\cite{lei2024mosca}                      &20.591      &0.777	&0.130	&- 		  &-      &92.74   &9.80   &102.54 \\
Playground & MoSca + Feature 3DGS~\cite{zhou2024feature}    &20.569	    &0.776	&0.124	&0.922	&0.447	&857.32  &61.38  &918.70      \\
Playground & Ours (single CLIP head)                        &20.463	    &0.775	&0.130	&0.922  &0.430	&141.02  &9.20   &150.22   \\
Playground & Ours (full model)                                          &20.536	    &0.775	&0.130	&0.913	&0.419	&140.44  &10.21  &150.65   \\
\midrule
Mean & MoSca~\cite{lei2024mosca}                      &25.166      &0.805	&0.111	&- 		  &-      &53.230   &14.496   &67.726 \\
Mean & MoSca + Feature 3DGS~\cite{zhou2024feature}    &25.191	    &0.806	&0.109	&0.881	&0.506	&491.979  &101.929  &593.907      \\
Mean & Ours (single CLIP head)                        &25.186	    &0.805	&0.112	&0.880  &0.510	&80.697   &14.597   &95.294   \\
Mean & Ours (full model)                                          &25.197	    &0.805	&0.110	&0.876	&0.503	&80.183   &15.274   &95.457   \\

\bottomrule
\end{tabular}
}
\end{center}
\caption{\textbf{Detailed Performance of 7 scenes from the NVIDIA Dataset.}}
\label{tab:nvidia_metrics}
\end{table*}

\section{Baseline Comparisons}
\label{sec:Baseline Comparisons}
\paragraph{Segment Anything (SAM2)}
In the main paper, we highlighted that our SAM2 inference implementation from the rendered unified latent feature map only needs to interact with the SAM2 decoding architecture. This is an improvement over the naive approach, which renders the RGB images (ordinary novel view synthesis like MoSca~\cite{lei2024mosca}) first and then processes it through the entire SAM2 encoder-decoder pipeline. 

We quantitatively evaluate our approach (denoted as Feature) against the naive baseline (rendered novel view RGB + SAM2, denoted as RGB) on two datasets (Nvidia~\cite{liu2023robust} and Nerfies~\cite{park2021nerfies}) in~\cref{tab:sam2}. A point prompt is randomly selected on the dynamic object in the first frame of the ground-truth novel view video, and SAM2 is used to generate per-frame ground-truth masks. We evaluate \textit{mIoU} across the entire dataset and repeat the experiments three times (Exp1-3) to assess generalizability. Our approach outperforms the baseline, which, though yielding smoother masks, lacks robustness. We observe that artifacts in RGB rendering can mislead SAM2 during encoding, causing ambiguity and highly inaccurate segmentation. In contrast, our feature space inference mitigates these issues, enhancing robustness and increasing speed.

\begin{figure*}[t]
  \includegraphics[width=\linewidth]{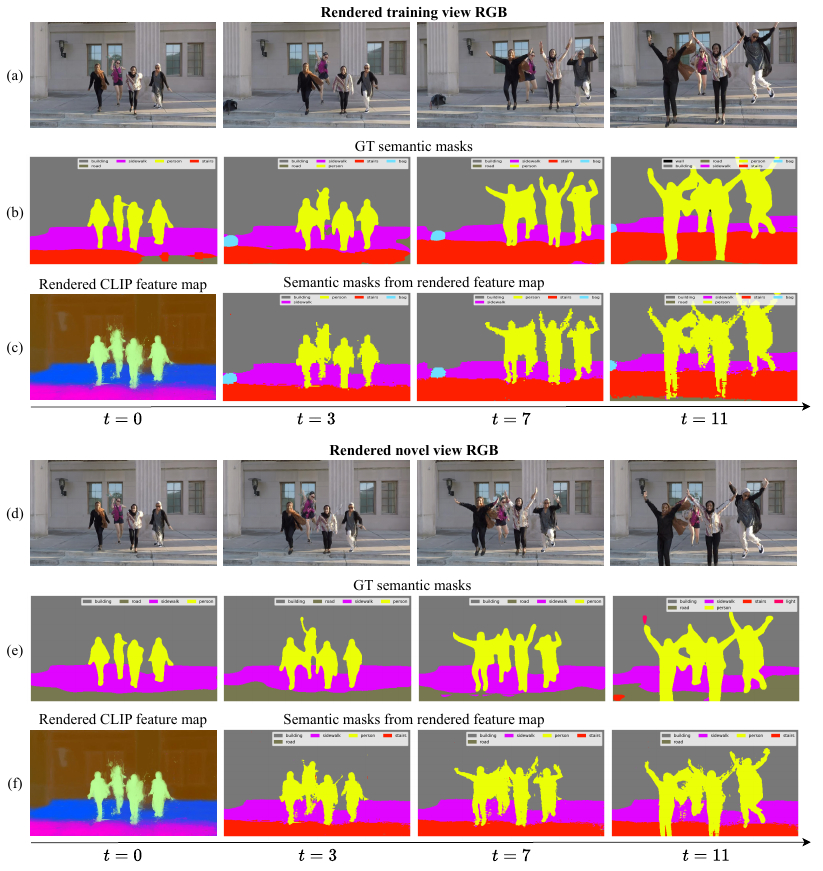}\vspace{-3mm}
  \caption{\textbf{CLIP semantic segmentation quality comparison}. We compare the CLIP semantic segmentation quality between ground-truth (inference from RGB) and our implementation (inference from feature map) for both training and novel views.} 
  \label{fig:clip_compare}
\end{figure*}

\paragraph{Semantic Segmentation (CLIP-Lseg)}
In main paper Tab. \textcolor{cvprblue}{1}, we provide the average performance metrics of our method and the comparison baselines across the 7 scenes of NVIDIA dataset. Here in~\cref{tab:nvidia_metrics}, we present the detailed performance metrics on each scene. 
We can conclude that our full model with a versatile feature field is not only fast and compact but also on par with any other task-specific feature field distillation methods. Additionally,~\cref{fig:clip_compare} presents the qualitative results of semantic segmentation on ``Jumping" scene using our full model, demonstrating the reliability of our approach.

We provide additional quantitative results on the Nerfies dataset~\cite{park2021nerfies} in~\cref{tab:lseg} to evaluate our method’s generalization capability. While we do not claim improving downstream task performance as a contribution, our method surpasses the naive baseline (novel view rendered per-frame RGB + LSeg) on this dataset while achieving 7.7$\times$ faster inference.

\section{Ablation Studies}
\label{sec:Ablation Studies}

\begin{figure*}[t]
  \centering
  \begin{minipage}{0.48\linewidth}
    \includegraphics[width=\linewidth]{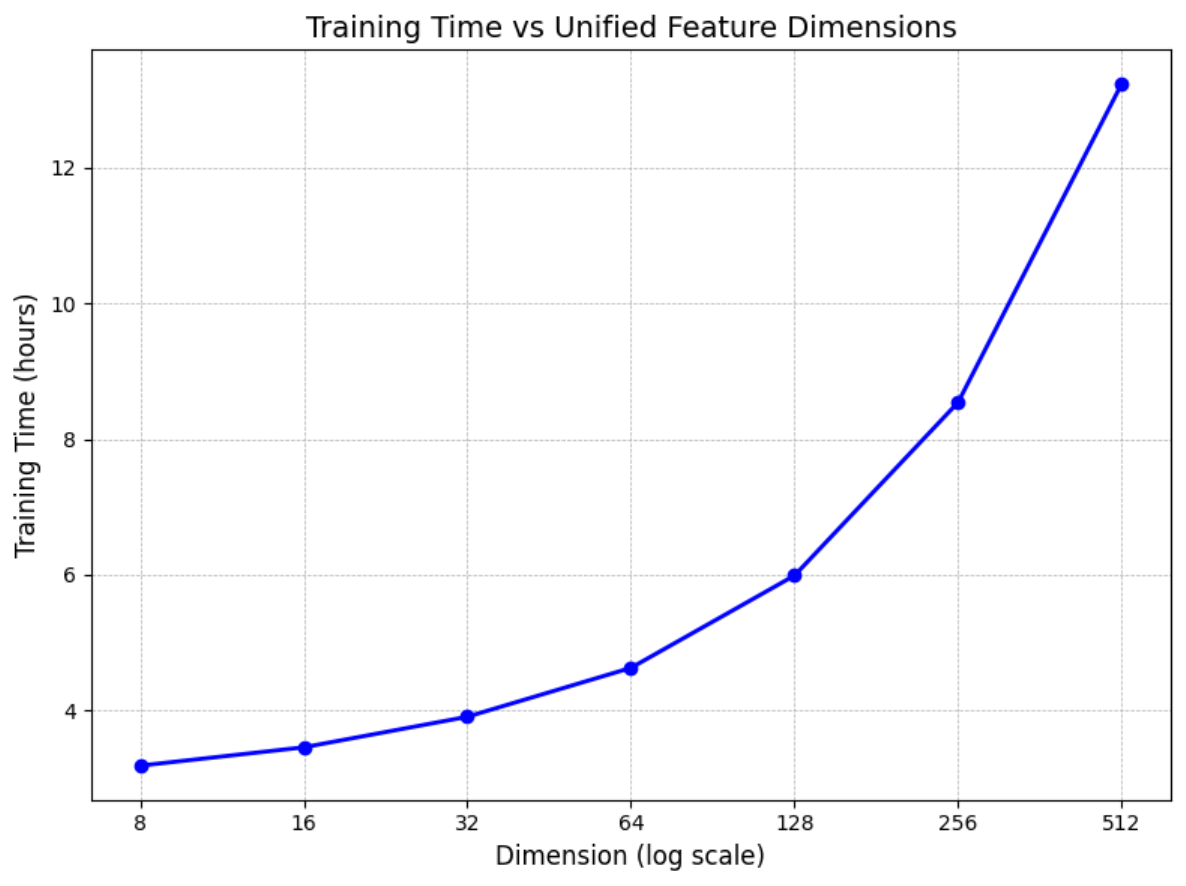}
    \caption{\textbf{Training Time vs Unified Latent Feature Dimensions} We show the training time required with different dimensions of unified latent feature map.}
    \label{fig:Train_time_vs_Dimension}
  \end{minipage}\hfill
    \begin{minipage}{0.48\linewidth}
    \includegraphics[width=\linewidth]{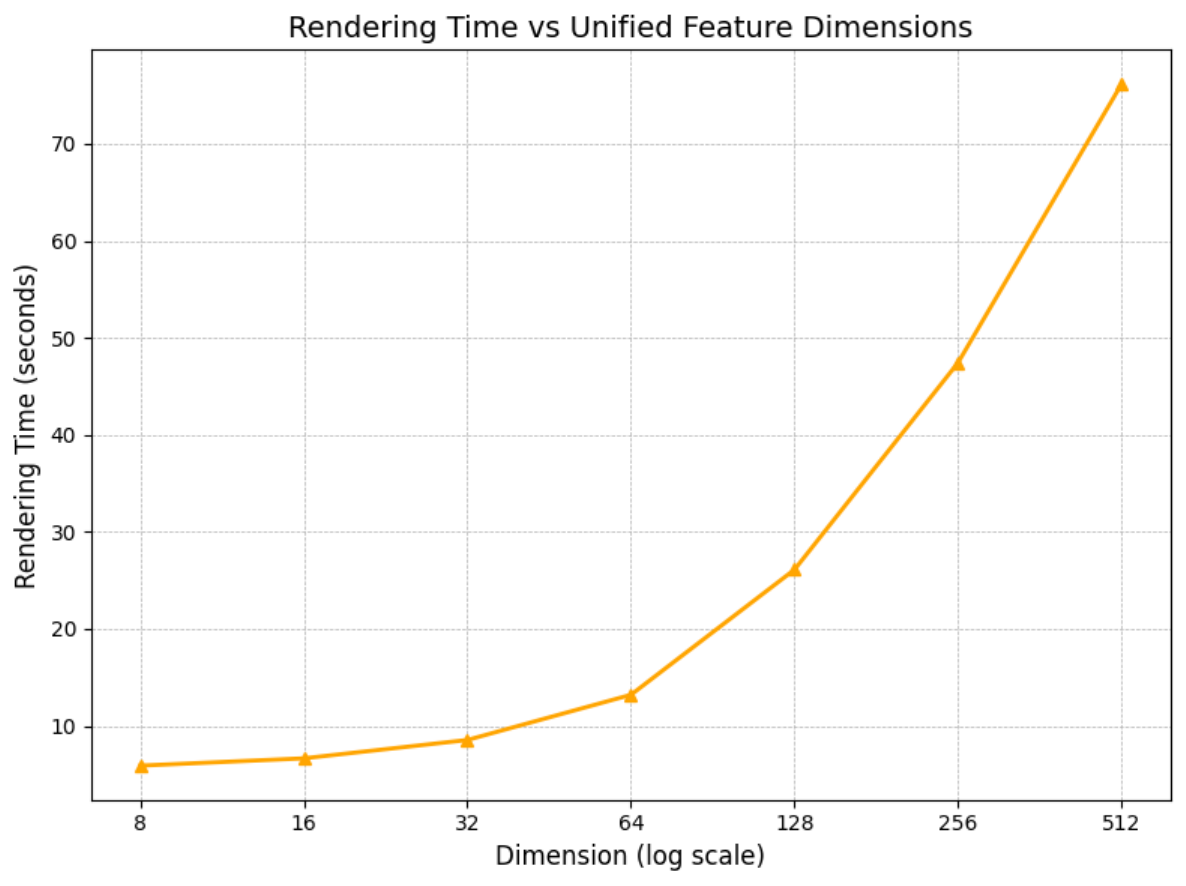}
    \caption{\textbf{Rendering Time vs Unified Latent Feature Dimensions} We show the rendering time required for different dimensions of unified latent feature map.}
    \label{fig:Render_time_vs_Dimension}
  \end{minipage}\hfill

  \begin{minipage}{0.48\linewidth}
    \includegraphics[width=\linewidth]{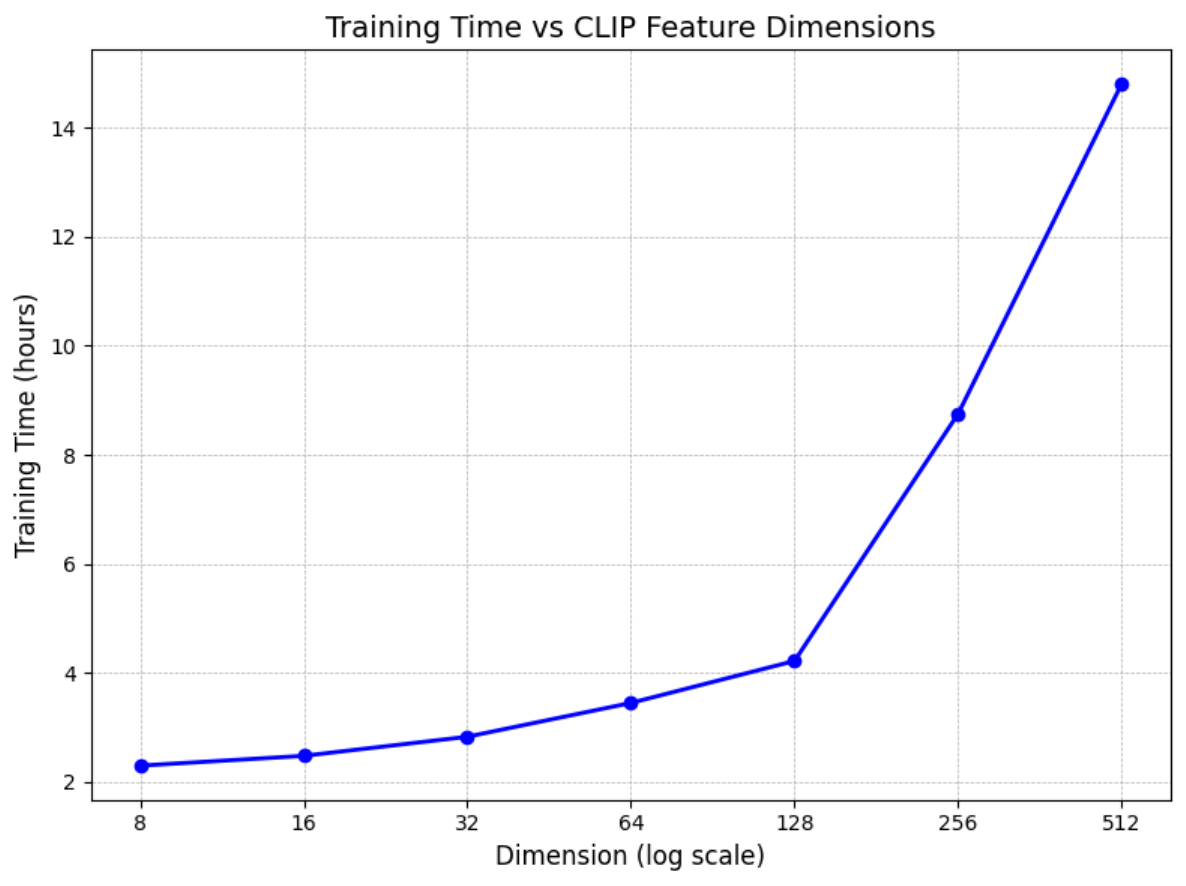}
    \caption{\textbf{Training Time vs CLIP Feature Dimensions} We show the training time required with different dimensions of rendered CLIP features.}
    \label{fig:clip_train_time_vs_Dimension}
  \end{minipage}\hfill 
    \begin{minipage}{0.48\linewidth}
    \includegraphics[width=\linewidth]{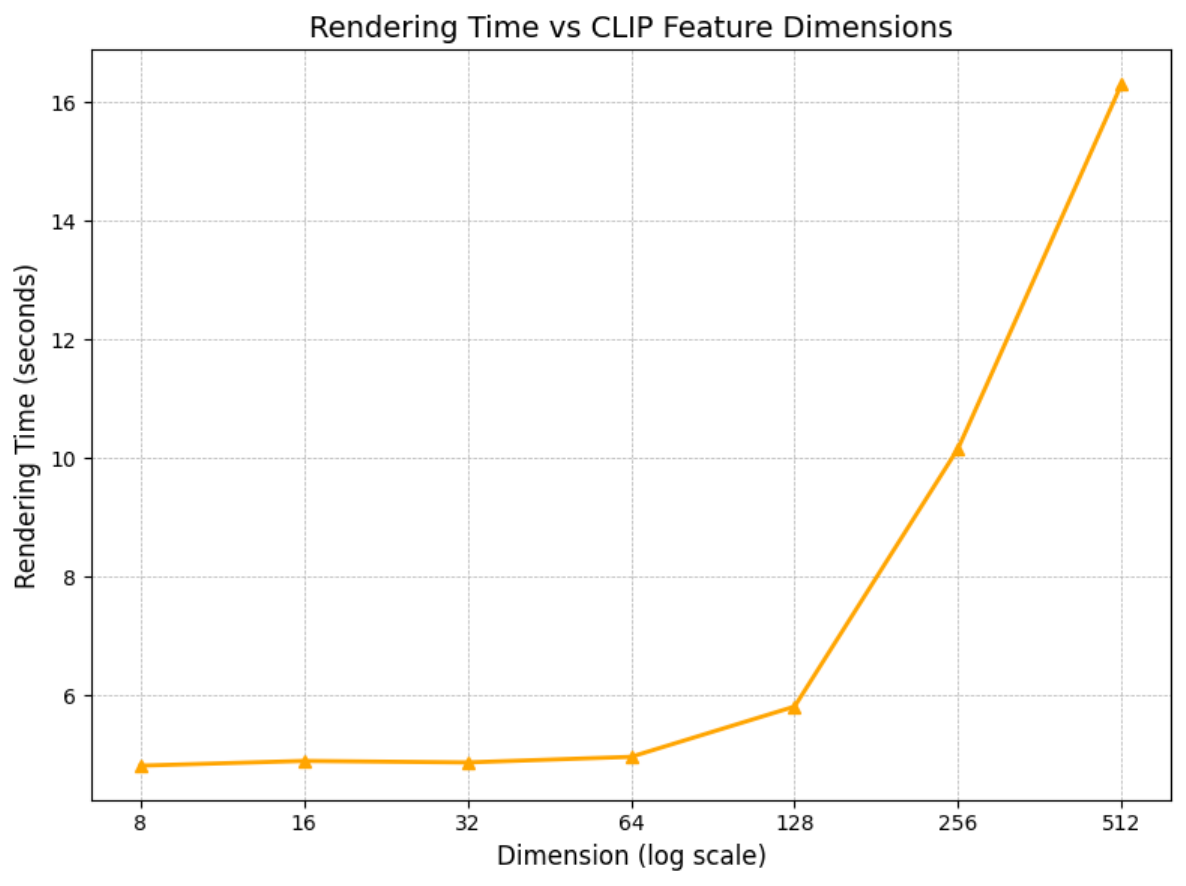}
    \caption{\textbf{Rendering Time vs CLIP Feature Dimensions} We show the rendering time required for different dimensions of rendered CLIP features.}
    \label{fig:clip_render_time_vs_Dimension}
  \end{minipage}\hfill
  \begin{minipage}{0.48\linewidth}
    \includegraphics[width=\linewidth]{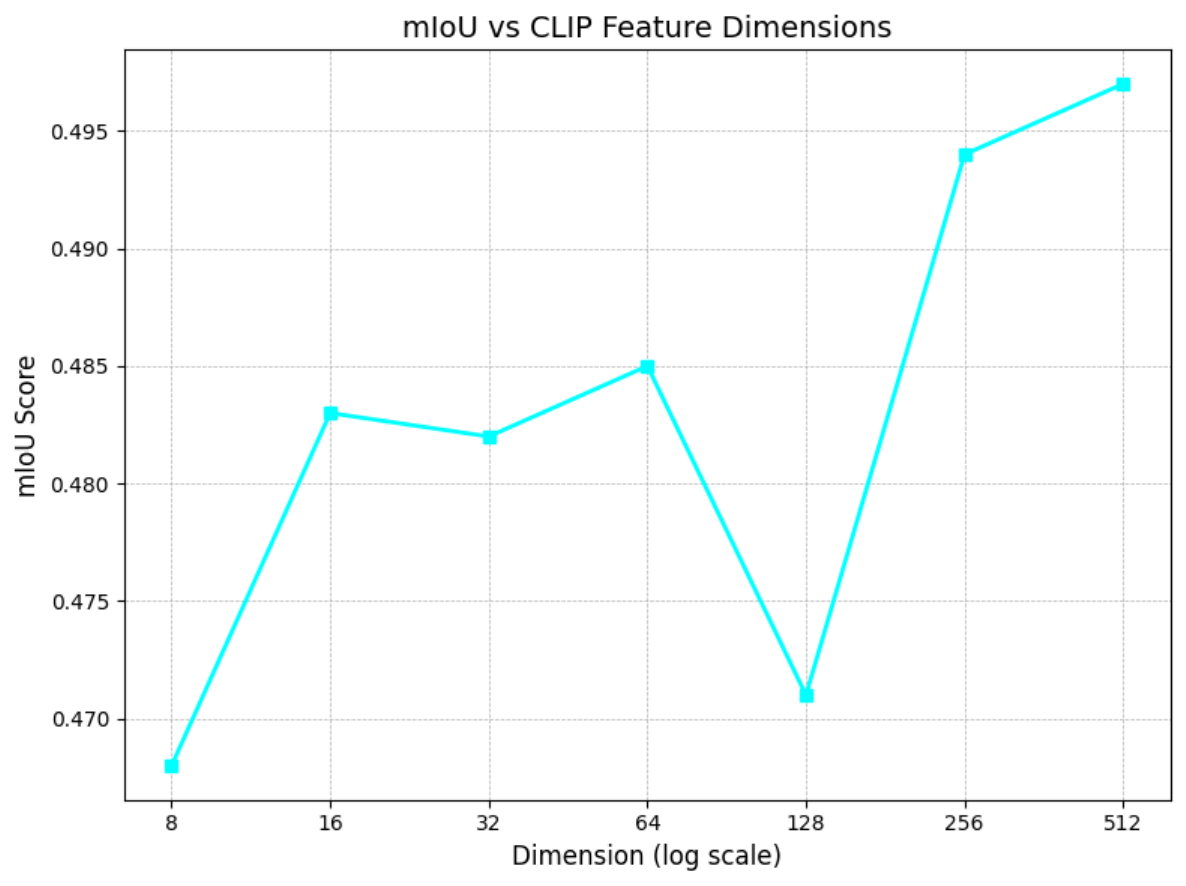}
    \caption{\textbf{mIoU vs CLIP Feature Dimensions} We show mIoU with respect to different rendered CLIP feature dimensions.}
    \label{fig:clip_mIoU_vs_Dimension}
  \end{minipage}\hfill
  \begin{minipage}{0.48\linewidth}
    \includegraphics[width=\linewidth]{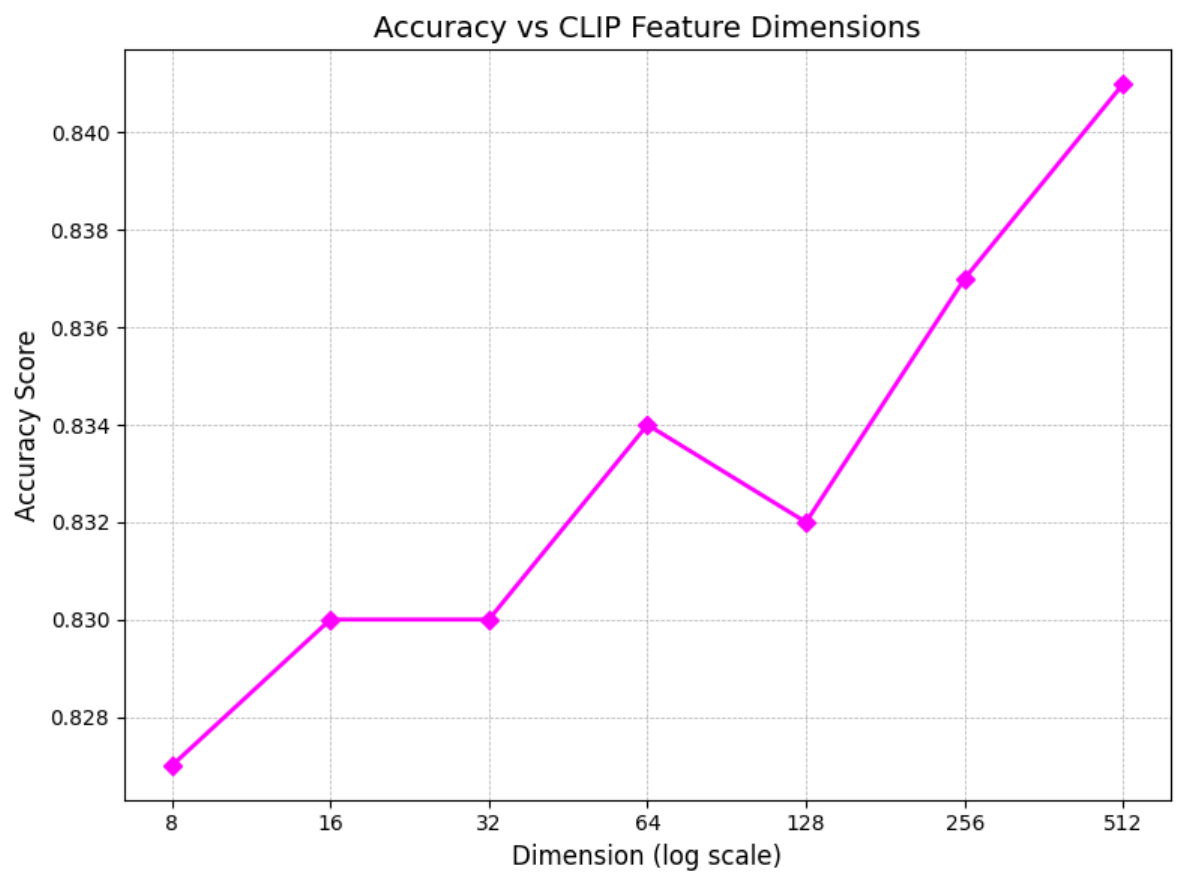}
    \caption{\textbf{Accuracy vs CLIP Feature Dimensions} We show accuracy with respect to different rendered CLIP feature dimensions.}
    \label{fig:clip_Accuracy_vs_Dimension}
  \end{minipage}\hfill
\end{figure*}

In all our experiments, we set the default dimension of the unified latent feature to 32, striking a balance between speed and quality. In this section, we present an ablation study to explore the impact of varying the feature dimensions. In ~\cref{fig:Train_time_vs_Dimension} and ~\cref{fig:Render_time_vs_Dimension} we compare the training and rendering time for different dimensions of our unified latent feature. Notably, a dimension of 32 marks the threshold where further increases in dimensions lead to a significant increase in training and rendering times. This makes larger dimensions impractical for our use case.

\begin{figure*}[t]
  \centering
  \includegraphics{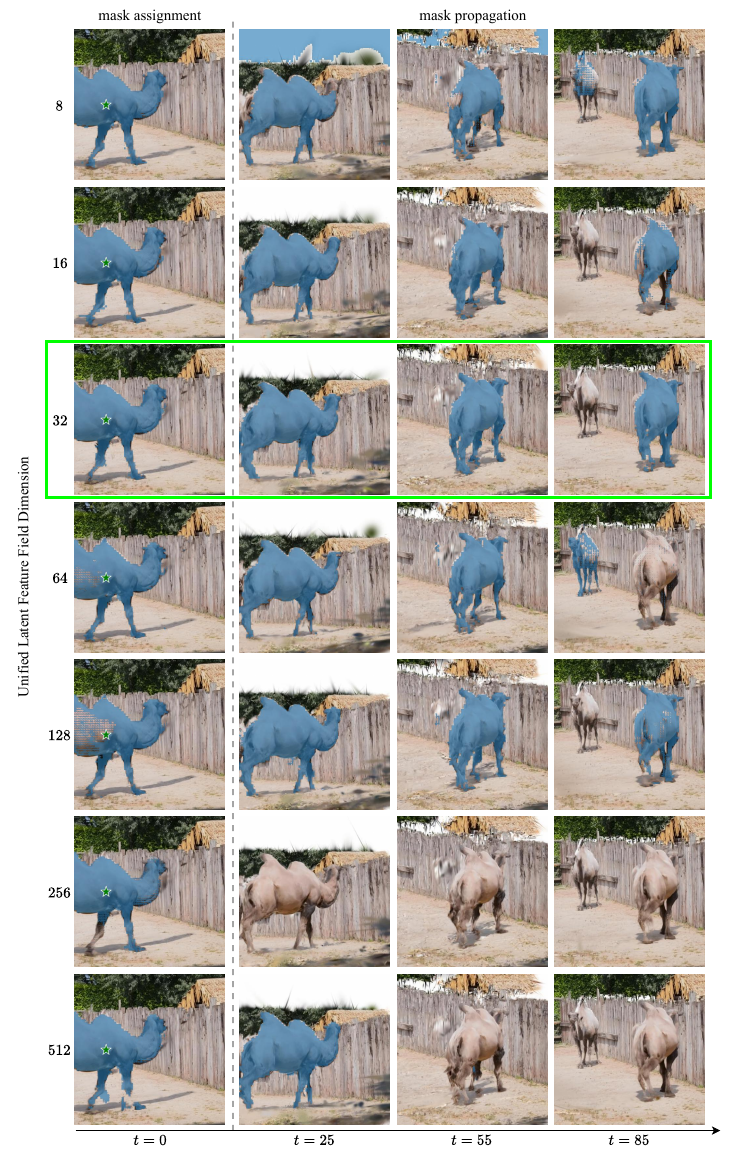}\vspace{-3mm}
  \caption{\textbf{SAM2 segmentation quality comparison for different dimensions of unified latent feature maps} Best performing SAM2 segmentation is derived from the 32-dimensional unified latent feature map.} 
  \label{fig:sam2_vs_dim}
\end{figure*}

\paragraph{Segment Anything (SAM2)}
In ~\cref{fig:sam2_vs_dim} we compare and contrast the SAM2 inference experiment results derived from different dimensions unified latent feature maps (8, 16, 32, 64, 128, 256, 512) to justify our choice of 32 dimensions. For dimensions higher than 32, we can see that the segmentation masks are of poor quality and cannot be accurately propagated through the video frames. For feature dimension 8, we see that while the segmentation mask tracks our intended object, it also erroneously includes much of the empty background. The segmentation masks for the 16 dimension experiment track accurately but do not fully cover the intended object at $t=55$ and $t=85$. Overall, the segmentation masks produced from 32 dimensional unified latent feature maps are the best in terms of segmentation quality and tracking accuracy. Given the relatively fast training and rendering, 32 should be the optimal dimension for our unified latent feature field in regards to SAM2 segmentation results.

\paragraph{Semantic Segmentation (CLIP-LSeg)}
We study the effect of different latent scene feature dimensions on the ``Jumping" scene of NVIDIA. In~\cref{tab:jumping_feat_metrics} (with plots~\cref{fig:clip_train_time_vs_Dimension} and~\cref{fig:clip_render_time_vs_Dimension}), we report the training time as well as the performance of the semantic segmentation. The results show that compared to 512, which is the original dimension of CLIP-LSeg feature, our method with $dim=32$ achieves comparable performance on mIoU and accuracy, while being $5.23 \times$ faster on training. We also report the image quality metrics in~\cref{tab:jumping_rgb_metrics} (with plots~\cref{fig:clip_mIoU_vs_Dimension} and~\cref{fig:clip_Accuracy_vs_Dimension}) which shows that our feature field distillation method does not affect the image quality. 

\begin{table*}[h]
    \vspace{-2mm}
    \centering
    \begin{tabular}{l cc cc cc}
        \toprule
        \multirow{2}{*}{\textbf{Nerfies} Scene} & \multicolumn{2}{c}{mIoU $\uparrow$} & \multicolumn{2}{c}{Accuracy $\uparrow$} & \multicolumn{2}{c}{Time (s) $\downarrow$} \\
        & RGB & Feature & RGB & Feature & RGB & Feature \\
        \midrule
        Broom    & 0.193 & \textbf{0.333} & 0.321 & \textbf{0.610} & 207.59 & \textbf{30.22} \\
        Curls    & \textbf{0.514} & 0.443 & \textbf{0.877} & 0.872 & 155.25 & \textbf{20.82} \\
        Tail     & 0.261 & \textbf{0.338} & 0.652 & \textbf{0.860} & 389.89 & \textbf{46.43} \\
        Toby-sit & \textbf{0.504} & 0.470 & \textbf{0.757} & 0.737 & 355.82 & \textbf{45.96} \\
        \midrule
        Mean     & 0.368 & \textbf{0.396} & 0.652 & \textbf{0.770} & 277.14 & \textbf{35.86} \\
        \bottomrule
    \end{tabular}
    \vspace{-3mm}
    \caption{\textbf{Semantic Segmentation Quantitative Results on Nerfies Dataset.}}
    \label{tab:lseg}
\end{table*}

\begin{table*}[t]
\begin{center}
\fontsize{6pt}{6pt}\selectfont
\resizebox{0.9\linewidth}{!}{
\begin{tabular}{lccccccc} 
\toprule
Dimension & 8 & 16 & 32 & 64 & 128 & 256 & 512 \\
\midrule
Training Time (h)     & 2.30 & 2.48  & 2.83 & 3.45 & 4.22 & 8.75 & 14.80 \\
Rendering Time (s)   & 4.818 & 4.898 & 4.872 & 4.967 & 5.815 & 10.167 & 16.313 \\
mIoU$\uparrow$      & 0.468 & 0.483 & 0.482 & 0.485 & 0.471 & 0.494 & \textbf{0.497} \\
Accuracy$\uparrow$  & 0.827 & 0.831 & 0.830 & 0.834 & 0.832 & 0.837 & \textbf{0.841} \\
\bottomrule
\end{tabular}}
\vspace{-3mm}
\end{center}
\caption{\textbf{Evaluation of Semantic Segmentation Performance On NVIDIA Jumping Scene Across Different Dimensions.} This table presents the Time, mIoU, and Accuracy corresponding to each dimension level.}
\label{tab:jumping_feat_metrics}
\end{table*}

\begin{table*}[h]
\begin{center}
\fontsize{8pt}{7pt}\selectfont
\resizebox{0.9\linewidth}{!}{
\begin{tabular}{lcccccccc} \toprule
Dimension & 8 & 16 & 32 & 64 & 128 & 256 & 512 & MoSca\\
\midrule
 PSNR$\uparrow$     & 24.466 & 24.595 & 24.616 & 24.623 & 24.585 & \textbf{24.629} & 24.491 &24.558      \\
 SSIM$\uparrow$     & 0.788 & 0.793   & 0.793 & 0.790 & 0.792 & \textbf{0.796} & 0.789      &0.792      \\
 LPIPS$\downarrow$  & 0.092 & 0.091   & 0.090 & 0.091 & 0.091 & 0.090 & \textbf{0.095}      &0.092      \\
\bottomrule
\end{tabular}}
\vspace{-3mm}
\end{center}
\caption{\textbf{Evaluation of Image Quality Metrics On NVIDIA Jumping Scene Across Different Dimensions.} This table presents the PSNR, SSIM, and LPIPS values corresponding to each dimension level.}
\label{tab:jumping_rgb_metrics}
\end{table*}




\clearpage




\end{document}